\documentclass[dvipsnames]{article}

\usepackage[final]{delta_tuning}

\usepackage{threeparttable}
\usepackage{colortbl} % 用于给单元格加背景色
\usepackage{xcolor}   % 定义颜色
\usepackage{booktabs} % 美化表格
\usepackage{amsmath}
\usepackage{siunitx} % For proper number alignment
\usepackage[utf8]{inputenc} % allow utf-8 input
\usepackage[T1]{fontenc}    % use 8-bit T1 fonts
\usepackage{url}            % simple URL typesetting
\usepackage{booktabs}       % professional-quality tables
\usepackage{amsfonts}       % blackboard math symbols
\usepackage{nicefrac}       % compact symbols for 1/2, etc.
\usepackage{booktabs} 
\usepackage{multirow}
\usepackage{wrapfig}
\usepackage{amssymb}
\usepackage{epigraph}
\usepackage{makecell}
\usepackage{listings}
\usepackage{pdflscape}
\usepackage[labelfont=bf]{caption} % For bold caption labels

\usepackage[ruled,vlined]{algorithm2e}
\usepackage{color, soul}
\usepackage{microtype}      % microtypography
\usepackage{xcolor}         % colors
\usepackage{geometry}
\usepackage{times}
\usepackage{ragged2e}
\usepackage{amsmath}
\usepackage{bm}
\usepackage{latexsym}
\usepackage{subfigure}
\usepackage{graphicx}
\usepackage{float}
\usepackage{longtable}
\usepackage{booktabs} 
\usepackage{multirow}
\usepackage{amssymb}
\usepackage{longtable}
\usepackage{tabu}
\usepackage{bbding}
\usepackage{awesomebox}
\usepackage{bbding}
\usepackage[most,breakable]{tcolorbox}
\usepackage{etoolbox}
\usepackage{fancyhdr}
\usepackage{lipsum}
\usepackage{CJKutf8}
\usepackage{pdfpages}
\usepackage{multicol}
\usepackage{pgffor}

\usepackage{caption} 
\usepackage{chngcntr}
\usepackage[colorlinks, linkcolor=RoyalBlue, anchorcolor=BrickRed, citecolor=RoyalBlue, urlcolor=RoyalBlue]{hyperref}

\usepackage{cleveref}
\crefname{section}{§}{§§}
\Crefname{section}{§}{§§}

\newcommand\ourmodel{UI-TARS\xspace}

\newcommand\refsec[1]{Section~\hyperref[sec:#1]{\ref{sec:#1}}}
\newcommand\refsecs[2]{\hyperref[sec:#1]{§\ref{sec:#1}:~\textsc{#1}}, \hyperref[sec:#2]{§\ref{sec:#2}:~\textsc{#2}}}

\usepackage[tikz]{bclogo}
\definecolor{msftBlue}{RGB}{0,164,239}
\definecolor{msftGreen}{RGB}{127,186,0}
\definecolor{msftYello}{RGB}{255,185,0}
\definecolor{msftBlack}{RGB}{0,0,0}

\newcommand\icon{\raisebox{-3.7pt}{\includegraphics[width=1.15em]{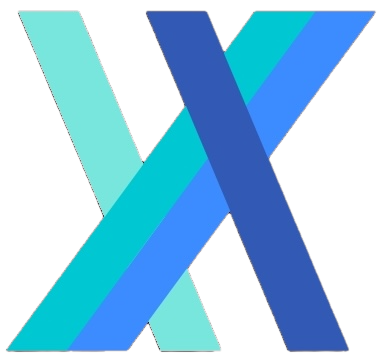}}}

\newtcolorbox{myboxnote}[1][]{
  breakable,
  title=#1,
  colback=cyan!0,
  colbacktitle=cyan!0,
  coltitle=black,
  fonttitle=\bfseries,
  bottomrule=0pt,
  toprule=0pt,
  leftrule=1.5pt,
  rightrule=1.5pt,
  titlerule=0pt,
  arc=0pt,
  outer arc=0pt,
  colframe=lightgray,
}
\usepackage{mdframed}

\definecolor{academicblue}{RGB}{54, 95, 145}

\newtcbox{\smybox}[1][red]{on line,
arc=1pt,colback=#1!10!white,colframe=#1!100!black,
before upper={\rule[-3pt]{0pt}{10pt}},
boxsep=0pt,left=6pt,right=6pt,top=2pt,bottom=0pt,boxrule=0pt,leftrule=1pt,rightrule=1pt}

\newenvironment{itemize*}%
 {\leftmargini=20pt\begin{itemize}%
  \setlength{\itemsep}{3pt}%
  \setlength{\parskip}{0pt}%
  }%
 {\end{itemize}}
\newenvironment{enumerate*}%
 {\begin{enumerate}%
  \setlength{\itemsep}{0pt}%
  \setlength{\parskip}{0pt}}%
 {\end{enumerate}}

\usepackage{fancyhdr} % to change header and footers
\usepackage{blindtext} % to quickly get a full document
\usepackage{ulem}

\usepackage{xparse}
\NewDocumentCommand{\heng}
{ mO{} }{\textcolor{red}{\textsuperscript{\textit{Heng}}\textsf{\textbf{\small[#1]}}}}

\usepackage{colortbl}

\title{
\icon\textsc{\ourmodel}: \\ Pioneering Automated GUI Interaction with Native Agents
}

\author{\small{Yujia Qin$^\dag\thanks{Indicates equal contribution. \ \ $^\dag$ Corresponding author. \ \ $^\diamondsuit$ The work is done when interning at ByteDance.}$\hspace{0.5em}, Yining Ye$^{*\diamondsuit}$, Junjie Fang$^{*}$, Haoming Wang$^{*}$, Shihao Liang$^{*}$, Shizuo Tian$^{\diamondsuit}$, Junda Zhang, Jiahao Li,} \\
\small{\textbf{Yunxin Li, Shijue Huang, Wanjun Zhong, Kuanye Li, Jiale Yang, Yu Miao, Woyu Lin, Longxiang Liu, Xu Jiang,}} \\
\small{\textbf{Qianli Ma, Jingyu Li, Xiaojun Xiao, Kai Cai, Chuang Li, Yaowei Zheng, Chaolin Jin, Chen Li, Xiao Zhou,}} \\ 
\small{\textbf{Minchao Wang, Haoli Chen, Zhaojian Li, Haihua Yang, Haifeng Liu, Feng Lin, Tao Peng, Xin Liu, Guang Shi$^{\dag}$}}\hspace{0.5em} \\
\small{ByteDance Seed, $^{\diamondsuit}$Tsinghua University} \\
\small{\texttt{\{yujia.qin, shiguang.sg\}@bytedance.com}}
}

\begin{document}

\maketitle

\begin{abstract}
This paper introduces \ourmodel, a native GUI agent model that solely perceives the screenshots as input and performs human-like interactions (e.g., keyboard and mouse operations). Unlike prevailing agent frameworks that depend on heavily wrapped commercial models (e.g., GPT-4o) with expert-crafted prompts and workflows, \ourmodel is an end-to-end \textbf{model} that outperforms these sophisticated \textbf{frameworks}. Experiments demonstrate its superior performance: \ourmodel achieves SOTA performance in 10+ GUI agent benchmarks evaluating perception, grounding, and GUI task execution (see below). Notably, in the OSWorld benchmark, \ourmodel achieves scores of $24.6$ with $50$ steps and $22.7$ with $15$ steps, outperforming Claude’s $22.0$ and $14.9$ respectively. In AndroidWorld, \ourmodel achieves $46.6$, surpassing GPT-4o’s $34.5$.
\ourmodel incorporates several key innovations: (1) \textbf{Enhanced Perception}: leveraging a large-scale dataset of GUI screenshots for context-aware understanding of UI elements and precise captioning; 
(2) \textbf{Unified Action Modeling}, which standardizes actions into a unified space across platforms and achieves precise grounding and interaction through large-scale action traces; (3) \textbf{System-2 Reasoning}, which incorporates deliberate reasoning into multi-step decision making,  involving multiple reasoning patterns such as task decomposition, reflection thinking, milestone recognition, etc. (4) \textbf{Iterative Training with Reflective Online Traces}, which addresses the data bottleneck by automatically collecting, filtering, and reflectively refining new interaction traces on hundreds of virtual machines. Through iterative training and reflection tuning, \ourmodel continuously learns from its mistakes and adapts to unforeseen situations with minimal human intervention. We also analyze the evolution path of GUI agents to guide the further development of this domain. \ourmodel is open sourced at \url{https://github.com/bytedance/UI-TARS}.

\begin{figure}[ht]
    \centering
    \begin{minipage}{0.95\textwidth}
        \centering
        \includegraphics[width=\textwidth]{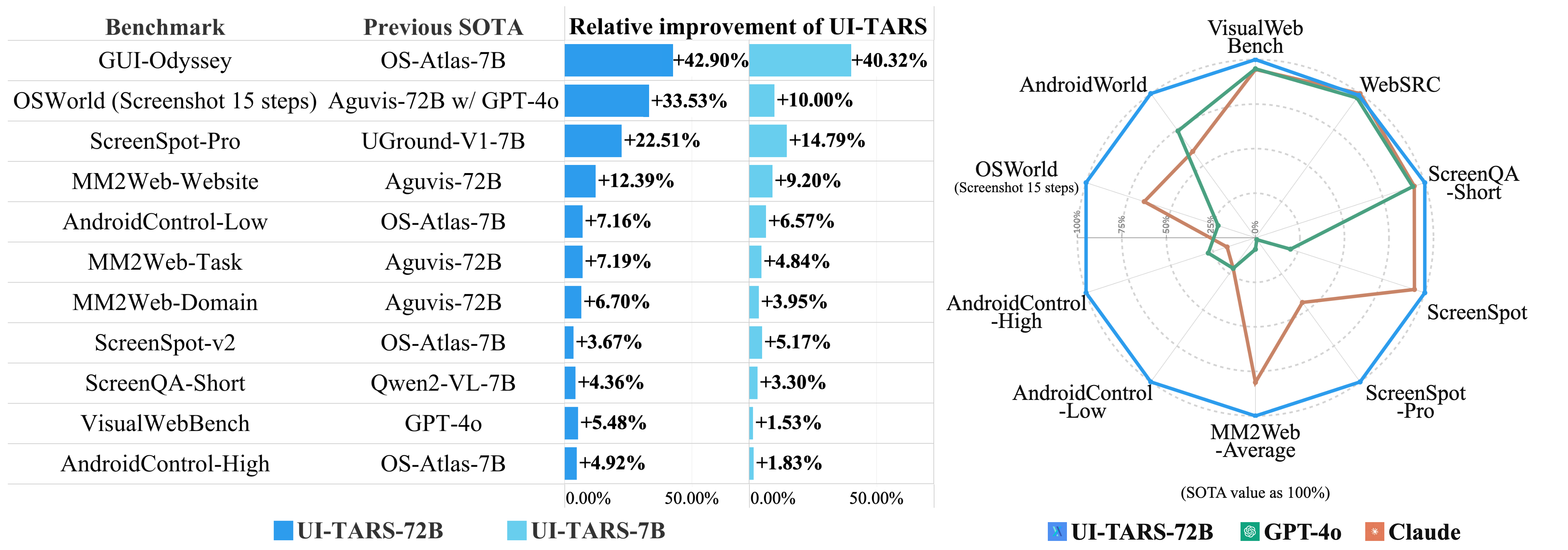}
        % \caption{MM2Web}
    \end{minipage}
    % \begin{minipage}{0.35\textwidth}
    %     \centering
    %     \includegraphics[width=\textwidth]{figures/Exp_UITARSvsAPI.jpeg}
    %     % \caption{Android Control}
    % \end{minipage}
\end{figure}
\end{abstract}

\newpage
{
  \hypersetup{linkcolor=RoyalBlue, linktoc=page}
  \tableofcontents
}

\newpage

\section{Introduction}
\label{sec:introduction}

\begin{figure}[!t]
    \centering
    \includegraphics[width=0.95\linewidth]{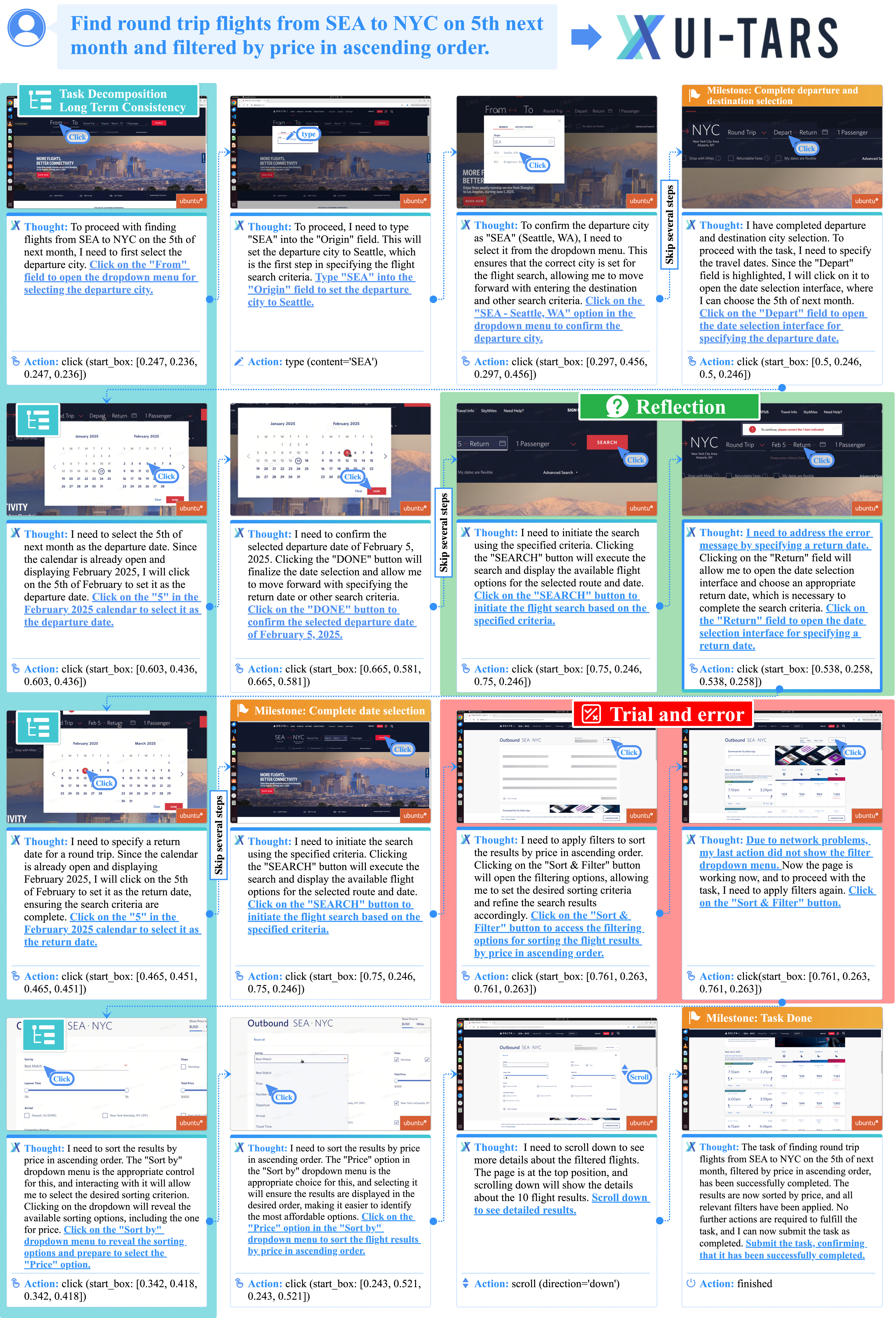}
    \caption{\small{A demo case of UI-TARS that helps user to find flights.}}
    \label{fig:case_intro}
\end{figure}

\textit{Autonomous agents}~\citep{wang2024survey,xi2023rise,10.1145/3704435} are envisioned to operate with minimal human oversight, perceiving their environment, making decisions, and executing actions to achieve specific goals. Among the many challenges in this domain, enabling agents to interact seamlessly with Graphical User Interfaces (GUIs) has emerged as a critical frontier~\citep{hu2024agents,zhang2024large,nguyen2024gui,wang2024gui,gao2024generalist}. GUI agents are designed to perform tasks within digital environments that rely heavily on graphical elements such as buttons, text boxes, and images. By leveraging advanced perception and reasoning capabilities, these agents hold the potential to revolutionize task automation, enhance accessibility, and streamline workflows across a wide range of applications.

The development of GUI agents has historically relied on hybrid approaches that combine textual representations (e.g., HTML structures and accessibility trees)~\citep{liu2018reinforcement,deng2023mind2webgeneralistagentweb,zhou2024webarenarealisticwebenvironment}. While these methods have driven significant progress, they suffer from limitations such as platform-specific inconsistencies, verbosity, and limited scalability~\citep{xu2024aguvis}. Textual-based methods often require system-level permissions to access underlying system information, such as HTML code, which further limits their applicability and generalizability across diverse environments. Another critical issue is that, many existing GUI systems follow an \textit{agent framework} paradigm~\citep{zhang2023appagent,wang2024mobile,wu2024copilot,zhang2024ufo,wang2024oscar,xie2024osworldbenchmarkingmultimodalagents}, where key functions are modularized across multiple components. These components often rely on specialized vision-language models (VLMs), e.g., GPT-4o~\citep{hurst2024gpt}, for understanding and reasoning~\citep{zhang2024ufo}, while grounding~\citep{lu2024omniparser} or memory~\citep{zhang2023appagent} modules are implemented through additional tools or scripts. Although this modular architecture facilitates rapid development in specific domain tasks, it relies on handcrafted approaches that depend on expert knowledge, modular components, and task-specific optimizations, which are less scalable and adaptive than end-to-end models. This makes the framework prone to failure when faced with unfamiliar tasks or dynamically changing environments~\citep{xia2024agentless}.

These challenges have prompted two key shifts towards \textit{native GUI agent model}: (1) the transition from \textbf{textual-dependent} to \textbf{pure-vision-based} GUI agents~\citep{fuyu-8b,hong2024cogagent}. ``Pure-vision'' means the model relies exclusively on screenshots of the interface as input, rather than textual descriptions (e.g., HTML). This bypasses the complexities and platform-specific limitations of textual representations, aligning more closely with human cognitive processes; and (2) the evolution from \textbf{modular agent frameworks} to \textbf{end-to-end agent models}~\citep{wu2024atlas,xu2024aguvis,qinghong2024showui,yang2024aria,claude-computer-use}. The end-to-end design unifies traditionally modularized components into a single architecture, enabling a smooth flow of information among modules. In philosophy, agent frameworks are \textbf{design-driven}, requiring extensive manual engineering and predefined workflows to maintain stability and prevent unexpected situations; while agent models are inherently \textbf{data-driven}, enabling them to learn and adapt through large-scale data and iterative feedback~\citep{putta2024agent}.

Despite their conceptual advantages, today's native GUI agent model often falls short in practical applications, causing their real-world impact to lag behind its hype. These limitations stem from two primary sources: (1) the GUI domain itself presents unique challenges that compound the difficulty of developing robust agents. (1.a) On the \textbf{perception} side, agents must not only recognize but also effectively interpret the high information-density of evolving user interfaces. (1.b) \textbf{Reasoning and planning} mechanisms are equally important in order to navigate, manipulate, and respond to these interfaces effectively. (1.c) These mechanisms must also leverage \textbf{memory}, considering past interactions and experiences to make informed decisions. (1.d) Beyond high-level decision-making, agents must also execute precise, low-level \textbf{actions}, such as outputting exact screen coordinates for clicks or drags and inputting text into the appropriate fields. (2) The transition from agent frameworks to agent models introduces a fundamental \textbf{data bottleneck}. Modular frameworks traditionally rely on separate datasets tailored to individual components. These datasets are relatively easy to curate since they address isolated functionalities. However, training an end-to-end agent model demands data that integrates all components in a unified workflow, capturing the seamless interplay between perception, reasoning, memory, and action. Such data, which comprise rich workflow knowledge from human experts, have been scarcely recorded historically. This lack of comprehensive, high-quality data limits the ability of native agents to generalize across diverse real-world scenarios, hindering their scalability and robustness.

To address these challenges, this paper focuses on advancing native GUI agent model. We begin by reviewing the evolution path for GUI agents (\cref{sec:roadmap}).
% and conducting a thorough literature review (\cref{sec:landscape}), 
By segmenting the development of GUI agents into key stages based on the degree of human intervention and generalization capabilities, we conduct a comprehensive literature review. Starting with traditional rule-based agents, we highlight the evolution from rigid, framework-based systems to adaptive native models that seamlessly integrate perception, reasoning, memory, and action. 
We also prospect the future potential of GUI agents capable of active and lifelong learning, which minimizes human intervention while maximizing generalization abilities.
To deepen understanding, we provide a detailed analysis of the core capabilities of the native agent model, which include: (1) perception, enabling real-time environmental understanding for improved situational awareness; (2) action, requiring the native agent model to accurately predict and ground actions within a predefined space; (3) reasoning, which emulates human thought processes and encompasses both System 1 and System 2 thinking; and (4) memory, which stores task-specific information, prior experiences, and background knowledge. We also summarize the main evaluation metrics and benchmarks for GUI agents.

Based on these analyses, we propose a native GUI agent model \ourmodel\footnote{\scriptsize{The name TARS is inspired by the character from \textcolor{blue}{\href{https://en.wikipedia.org/wiki/Interstellar_(film)}{\textit{Interstellar}}}, a witty and adaptable robotic crew member known for his problem-solving abilities and dynamic decision-making.}}, with a demo case illustrated in Figure~\ref{fig:case_intro}. \ourmodel incorporates the following core contributions:
\begin{itemize*}
    \item \textbf{Enhanced Perception for GUI Screenshots} (\cref{sec:perception}): GUI environments, with their high information density, intricate layouts, and diverse styles, demand robust perception capabilities. We curate a large-scale dataset by collecting screenshots using specialized parsing tools to extract metadata such as element types, bounding boxes, and text content from websites, applications, and operating systems. The dataset targets the following tasks: (1) element description, which provides fine-grained, structured descriptions of GUI components; (2) dense captioning, aimed at holistic interface understanding by describing the entire GUI layout, including spatial relationships, hierarchical structures, and interactions among elements; (3) state transition captioning, which captures subtle visual changes in the screen; (4) question answering, designed to enhance the agent's capacity for visual reasoning; and (5) set-of-mark prompting, which uses visual markers to associate GUI elements with specific spatial and functional contexts. These carefully designed tasks collectively enable \ourmodel to recognize and understand GUI elements with exceptional precision, providing a robust foundation for further reasoning and action.
    \item \textbf{Unified Action Modeling for Multi-step Execution} (\cref{sec:action}): we design a unified action space to standardize semantically equivalent actions across platforms. To improve multi-step execution, we create a large-scale dataset of action traces, combining our annotated trajectories and standardized open-source data. The grounding ability, which involves accurately locating and interacting with specific GUI elements, is improved by curating a vast dataset that pairs element descriptions with their spatial coordinates. This data enables \ourmodel to achieve precise and reliable interactions.
    \item \textbf{System-2 Reasoning for Deliberate Decision-making} (\cref{sec:reasoning}): robust performance in dynamic environments demands advanced reasoning capabilities. To enrich reasoning ability, we crawl $6$M GUI tutorials, meticulously filtered and refined to provide GUI knowledge for logical decision-making. Building on this foundation, we augment reasoning for all the collected action traces by injecting diverse reasoning patterns—such as task decomposition, long-term consistency, milestone recognition, trial\&error, and reflection—into the model. \ourmodel integrates these capabilities by generating explicit ``thoughts'' before each action, bridging perception and action with deliberate decision-making.
    \item \textbf{Iterative Refinement by Learning from Prior Experience} (\cref{sec:memory}): a significant challenge in GUI agent development lies in the scarcity of large-scale, high-quality action traces for training. To overcome this data bottleneck, \ourmodel employs an iterative improvement framework that dynamically collects and refines new interaction traces. Leveraging hundreds of virtual machines, \ourmodel explores diverse real-world tasks based on constructed instructions and generates numerous traces. Rigorous multi-stage filtering—incorporating rule-based heuristics, VLM scoring, and human review—ensures trace quality. These refined traces are then fed back into the model, enabling continuous, iterative enhancement of the agent's performance across successive cycles of training. Another central component of this online bootstrapping process is \textbf{reflection tuning}, where the agent learns to identify and recover from errors by analyzing its own suboptimal actions. We annotate two types of data for this process: (1) \textbf{error correction}, where annotators pinpoint mistakes in agent-generated traces and label the corrective actions, and (2) \textbf{post-reflection}, where annotators simulate recovery steps, demonstrating how the agent should realign task progress after an error. These two types of data create paired samples, which are used to train the model using Direct Preference Optimization (DPO)~\citep{rafailov2024direct}. This strategy ensures that the agent not only learns to avoid errors but also adapts dynamically when they occur. Together, these strategies enable \ourmodel to achieve robust, scalable learning with minimal human oversight.
\end{itemize*}

We continually train Qwen-2-VL 7B and 72B~\citep{wang2024qwen2vlenhancingvisionlanguagemodels} on approximately \textbf{50 billion} tokens to develop \ourmodel-7B and \ourmodel-72B. Through extensive experiments, we draw the following conclusions:

\begin{itemize*}
    \item \textbf{Overall Performance}: \ourmodel demonstrates SOTA performance across $10+$ GUI Agent benchmarks, covering evaluation for perception, grounding, and agent task execution. These results validate the effectiveness of our method, significantly outperforming competitive baselines such as GPT-4o and Claude Computer Use~\citep{claude-computer-use} in reasoning-intensive and dynamic scenarios.

    \item \textbf{Perception}: \ourmodel excels in GUI perception, effectively handling high information density and intricate layouts. Experiments confirm its ability to extract precise metadata, describe GUI elements, and generate detailed, context-aware captions. For example, \ourmodel-72B scored $82.8$ in VisualWebBench~\citep{liu2024visualwebbench}, higher than GPT-4o ($78.5$).

    \item \textbf{Grounding}: \ourmodel achieves high-precision grounding across mobile, desktop, and web environments by accurately associating GUI elements with their spatial coordinates. For example, it scores $38.1$ (SOTA) on a recently released challenging benchmark ScreenSpot Pro~\citep{li2024screenspot-pro}.
    % significantly outperforming OS-Atlas~\citep{wu2024atlas} (18.9) and UGround~\citep{gou2024navigatingdigitalworldhumans} (31.1).

    \item \textbf{Agent Capabilities}: extensive evaluations highlight \ourmodel’s superior agent capabilities. Experiments demonstrate exceptional performance in reasoning-intensive benchmarks, with the 72B variant particularly excelling in multistep and dynamic tasks. Notably, \ourmodel achieves excellent results on challenging benchmarks such as OSWorld~\citep{xie2024osworldbenchmarkingmultimodalagents} and AndroidWorld~\citep{rawles2024androidworlddynamicbenchmarkingenvironment}. In OSWorld, \ourmodel-72B achieves scores of $24.6$ with 50 steps and $22.7$ with 15 steps, outperforming Claude’s $22.0$ and $14.9$ respectively. In AndroidWorld, it achieves a score of $46.6$, outshining GPT-4o's $34.5$, further emphasizing its ability to tackle high-complexity real-world scenarios.
\end{itemize*}
\section{Evolution Path of GUI Agents}
\label{sec:roadmap}

GUI agents are particularly significant in the context of automating workflows, where they help streamline repetitive tasks, reduce human effort, and enhance productivity.
At their core, GUI agents are designed to facilitate the interaction between humans and machines, simplifying the execution of tasks. Their evolution reflects a progression from rigid, human-defined heuristics to increasingly autonomous systems that can adapt, learn, and even independently identify tasks. In this context, the role of GUI agents has shifted from simple automation to full-fledged, self-improving agents that increasingly integrate with the human workflow, acting not just as tools, but as collaborators in the task execution process.

\begin{figure}[!t]
    \centering
    \includegraphics[width=\linewidth]{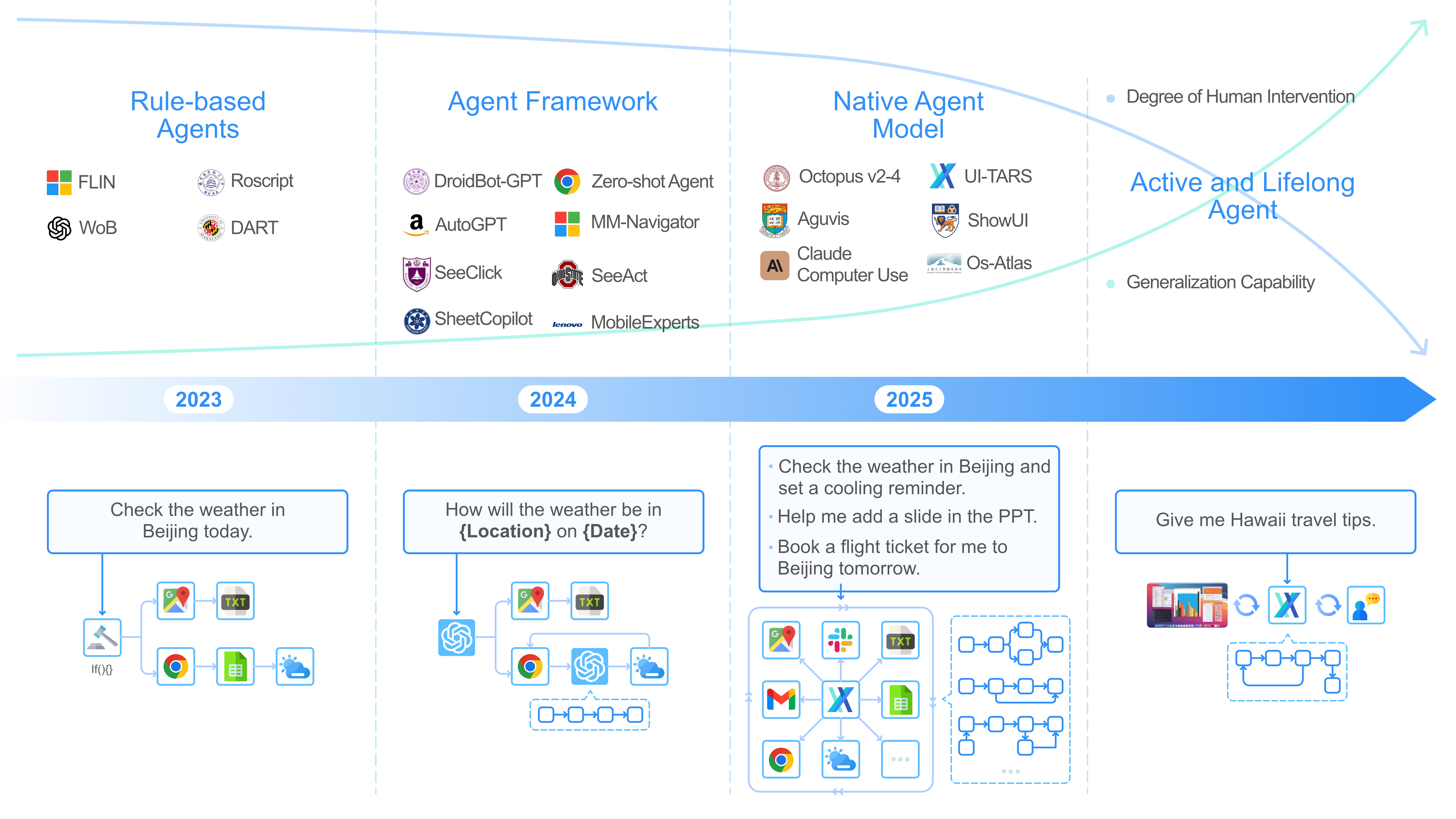}
    \caption{\small{The evolution path for GUI agents.}}
    \label{roadmap}
\end{figure}

Over the years, agents have progressed from basic rule-based automation to an advanced, highly automated, and flexible system that increasingly mirrors human-like behavior and requires minimal human intervention to perform its tasks. As illustrated in Figure~\ref{roadmap}, the development of GUI agents can be broken down into several key stages, each representing a leap in autonomy, flexibility, and generalization ability. Each stage is characterized by how much human intervention is required in the workflow design and learning process. 

% \subsection{Rule-based and API-driven Agents}
\subsection{Rule-based Agents}
\paragraph{Stage 1: Rule-based Agents}
In the initial stage, agents such as Robotic Process Automation (RPA) systems~\citep{10.1145/3511667,RePEc:spr:elmark:v:30:y:2020:i:1:d:10.1007_s12525-019-00365-8} were designed to replicate human actions in highly structured environments, often interacting with GUIs and enterprise software systems. These agents typically processed user instructions by matching them to predefined rules and invoking APIs accordingly. Although effective for well-defined and repetitive tasks, these systems were constrained by their reliance on human-defined heuristics and explicit instructions, hindering their ability to handle novel and complex scenarios.  At this stage, the agent cannot learn from its environment or previous experiences, and any changes to the workflow require human intervention.
Moreover, these agents require direct access to APIs or underlying system permissions, as demonstrated by systems like DART~\citep{1235451}, WoB~\citep{pmlr-v70-shi17a}, Roscript~\citep{9283959} and FLIN~\citep{mazumder-riva-2021-flin}. This makes it unsuitable for cases where such access is restricted or unavailable. This inherent rigidity constrained their applicability to scale across diverse environments.

The limitations of rule-based agents underscore the importance of transitioning to GUI-based agents that rely on visual information and explicit operation on GUIs instead of requiring low-level access to systems. Through visual interaction with interfaces, GUI agents unlock greater flexibility and adaptability, significantly expanding the range of tasks they can accomplish without being limited by predefined rules or the need for explicit system access. This paradigm shift opens pathways for agents to interact with unfamiliar or newly developed interfaces autonomously.

\subsection{From Modular Agent Framework to Native Agent Model}

Agent frameworks leveraging the power of large models (M)LLMs have surged in popularity recently. This surge is driven by the foundation models' ability to deeply comprehend diverse data types and generate relevant outputs via multi-step reasoning.  Unlike rule-based agents, which necessitate handcrafted rules for each specific task, foundation models can generalize across different environments and effectively handle tasks by interacting multiple times with environments. This eliminates the need for humans to painstakingly define rules for every new scenario, significantly simplifying agent development and deployment.

\paragraph{Stage 2: Agent Framework} Specifically, these agent systems mainly leverage the understanding and reasoning capabilities of advanced foundation models (e.g., GPT-4~\citep{openai2023gpt4} and GPT-4o~\citep{hurst2024gpt}) to enhance task execution flexibility, which become more flexible, framework-based agents. Early efforts primarily focused on tasks such as calling specific APIs or executing code snippets within text-based interfaces~\citep{wang2023enabling,li2024sheetcopilot,li2023zero,wen2023droidbot,nakano2021webgpt}. These agents marked a significant advancement from purely rule-based systems by enabling more automatic and flexible interactions. Autonomous frameworks like AutoGPT~\citep{yang2023auto} and LangChain allow agents to integrate multiple external tools, APIs, and services, enabling a more dynamic and adaptable workflow.

Enhancing the performance of foundation model-based agent frameworks often involves designing task-specific workflows and optimizing prompts for each component. For instance, some approaches augment these frameworks with specialized modules, such as short- or long-term memory, to provide task-specific knowledge or store operational experience for self-improvement. Cradle~\citep{tan2024towards} enhances foundational agents' multitasking capabilities by storing and leveraging task execution experiences. Similarly, \citet{song2024beyond} propose a framework for API-driven web agents that utilizes task-specific background knowledge to execute complex web operations. The Agent Workflow Memory (AWM) module~\citep{wang2024agent} further optimizes memory management by selectively providing relevant workflows to guide the agent's subsequent actions.
Another common strategy to improve task success is the incorporation of reflection-based, multi-step reasoning to refine action planning and execution. The widely recognized ReAct framework~\citep{yao2022react} integrates reasoning with the outcomes of actions, enabling more dynamic and adaptable planning. For multimodal tasks, MMNavigator~\citep{yan2023gpt4vwonderlandlargemultimodal} leverages summarized contextual actions and mark tags to generate accurate, executable actions. SeeAct~\citep{zheng2024gpt4visiongeneralistwebagent} takes a different approach by explicitly instructing GPT-4V to mimic human browsing behavior, taking into account the task, webpage content, and previous actions. 
Furthermore, multi-agent collaboration has emerged as a powerful technique for boosting task completion rates. MobileExperts~\citep{zhang2024mobileexperts}, for example, addresses the unique challenges of mobile environments by incorporating tool formulation and fostering collaboration among multiple agents. In summary, current advancements in agent frameworks heavily rely on optimizing plan and action generation through prompt engineering, centered around the capabilities of the underlying foundation models, ultimately leading to improved task completion.

\paragraph{Key Limitations of Agent Frameworks} Despite greater adaptability compared to rule-based systems, agent frameworks still rely on human-defined workflows to structure their actions. The ``agentic workflow knowledge''~\citep{wang2024agent} is manually encoded through custom prompts, external scripts, or tool-usage heuristics. This externalization of knowledge yields several drawbacks:
\begin{itemize*}
    \item \textbf{Fragility and Maintenance Overhead}: whenever tasks, interfaces, or usage scenarios evolve, the workflow’s manual rules or prompts must be re-crafted or extended by developers—an error-prone and labor-intensive process.
    \item \textbf{Disjoint Learning Paradigms}: framework-based methods rarely integrate new experience data to update the underlying LLM/VLM parameters. Instead, they rely on offline prompt-engineering or workflow design. As tasks deviate from the original domain, these frameworks often fail, limiting adaptability.
    \item \textbf{Module Incompatibility}: complex tasks demand multiple modules (e.g., visual parsing, memory stores, long-horizon planning) that must coordinate via prompts or bridging code. Inconsistencies or errors in any module can derail the entire pipeline, and diagnosing these issues typically requires domain experts to debug the flow.
\end{itemize*}
Thus, while agent frameworks offer quick demonstrations and are flexible within a narrow scope, they ultimately remain brittle when deployed in real-world scenarios, where tasks and interfaces continuously evolve. \textbf{This reliance on pre-programmed workflows, driven by human expertise, makes frameworks inherently non-scalable}. They depend on the foresight of developers to anticipate all future variations, which limits their capacity to handle unforeseen changes or learn autonomously. Frameworks are \textbf{design-driven}, meaning they lack the ability to learn and generalize across tasks without continuous human involvement.

\paragraph{Stage 3: Native Agent Model} 
In contrast, the future of autonomous agent development lies in the creation of native agent models, where workflow knowledge is embedded directly within the agent’s model through orientational learning. In this paradigm, tasks are learned and executed in an end-to-end manner, unifying perception, reasoning, memory, and action within a single, continuously evolving model. This approach is fundamentally \textbf{data-driven}, allowing for the seamless adaptation of agents to new tasks, interfaces, or user needs without relying on manually crafted prompts or predefined rules. Native agents offer several distinct advantages that contribute to their scalability and adaptability:
\begin{itemize*}
    \item \textbf{Holistic Learning and Adaptation}: because the agent’s policy is learned end-to-end, it can unify knowledge from perception, reasoning, memory, and action in its internal parameters. As new data or user demonstrations become available, the entire system (rather than just a single module or prompt) updates its knowledge. This empowers the model to adapt more seamlessly to changing tasks, interfaces, or user demands.
    \item \textbf{Reduced Human Engineering}: instead of carefully scripting how the LLM/VLM should be invoked at each node, native models learn task-relevant workflows from large-scale demonstrations or online experiences. The burden of “hardwiring a workflow” is replaced by data-driven learning. This significantly reduces the need for domain experts to handcraft heuristics whenever the environment evolves.
    \item \textbf{Strong Generalization via Unified Parameters}: Although manual prompt engineering can make the model adaptable to user-defined new tools, the model itself cannot evolve. Under one parameterized policy and a unified data construction and training pipeline, knowledge among environments like certain app features, navigation strategies, or UI patterns can be transferred across tasks, equipping it with strong generalization.
    \item \textbf{Continuous Self-Improvement}: native agent models lend themselves naturally to online or lifelong learning paradigms. By deploying the agent in real-world GUI environments and collecting new interaction data, the model can be fine-tuned or further trained to handle novel challenges.
\end{itemize*}
This data-driven, learning-oriented approach stands in contrast to the design-driven, static nature of agent frameworks. As for now, the development of GUI agent gradually reached this stage, which representative works like Claude Computer-Use~\citep{claude-computer-use}, Aguvis~\citep{xu2024aguvis}, ShowUI~\citep{qinghong2024showui}, OS-Atlas~\citep{wu2024atlas}, Octopus v2-4~\citep{chen2024octopusv2ondevicelanguage},
etc. These models mainly utilize existing world data to tailor large VLMs specifically for the domain of GUI interaction. 

\subsection{Active and Lifelong Agent (Prospect)}
\paragraph{Stage 4: Action and Lifelong Agent}
Despite improvements in adaptability, native agents still rely heavily on human experts for data labeling and training guidance. This dependence inherently restricts their capabilities, making them contingent upon the quality and breadth of human-provided data and knowledge.

The transition towards active and lifelong learning~\citep{sur2022system,ramamoorthy2024distributed} represents a crucial next step in the evolution of GUI agents. In this paradigm, agents actively engage with their environment to propose tasks, execute them, and evaluate the outcomes. These agents can autonomously assign self-rewards based on the success of their actions, reinforcing positive behaviors and progressively refining their capabilities through continuous feedback loops. This process of self-directed exploration and learning allows the agent to discover new knowledge, improve task execution, and enhance problem-solving strategies without heavy reliance on manual annotations or explicit external guidance.

These agents develop and modify their skills iteratively, much like continual learning in robotics~\citep{ayub2024interactive,soltoggio2024collective}, where they can learn from both successes and failures, progressively enhancing their generalization across an increasingly broad range of tasks and scenarios. The key distinction between native agent models and active lifelong learners lies in the autonomy of the learning process: native agents still depend on humans, whereas active agents drive their own learning by identifying gaps in their knowledge and filling them through self-initiated exploration.

In this work, we focus on building a scalable and data-driven native agent model, which paves the way for this active and lifelong agent stage. We begin by exploring the core capabilities necessary for such a framework (\cref{sec:landscape_core_ability}) and then introduce UI-TARS, our instantiation of this approach (\cref{sec:method}).
\begin{figure}[!t]
    \centering
    \includegraphics[width=\linewidth]{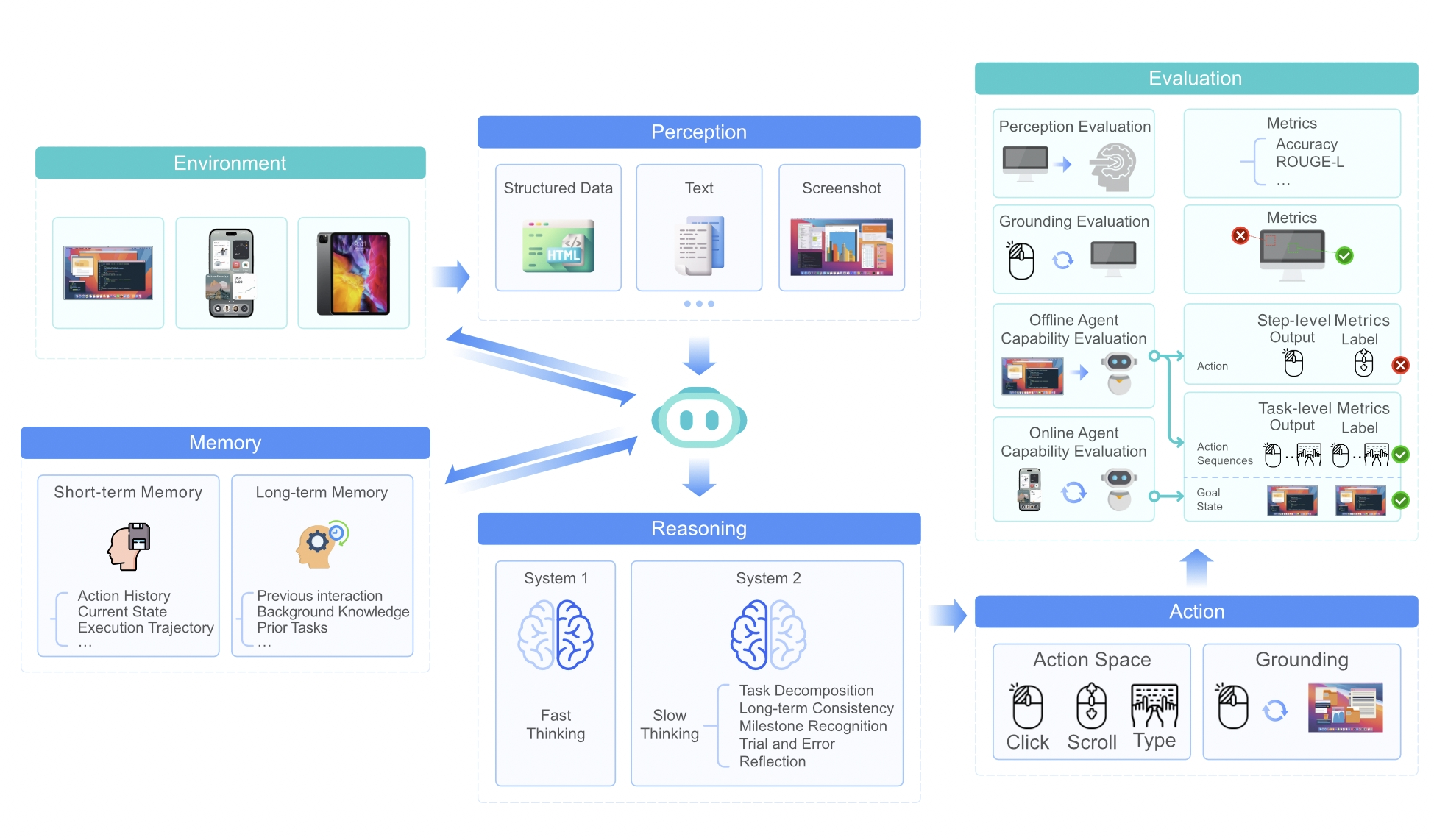}
    \caption{\small{An overview of core capabilities and evaluation for GUI agents.}}
    \label{core_cap}
\end{figure}

\section{Core Capabilities of Native Agent Model}
\label{sec:landscape_core_ability}

The native agent model internalizes modularized components from the previous agent framework into several core capabilities, thereby transitioning towards an end-to-end structure. 
% To gain a deeper understanding of the native agent model, 
To get a more profound understanding of the native agent model,
this section delves into an in-depth analysis of its core capabilities and reviews the current evaluation metrics and benchmarks.

\subsection{Core Capabilities}
As illustrated in Figure~\ref{core_cap}, our analysis is structured around four main aspects: perception, action, reasoning (system-1\&2 thinking), and memory.

\paragraph{Perception}
A fundamental aspect of effective GUI agents lies in their capacity to precisely perceive and interpret graphical user interfaces in real-time.
This involves not only understanding static screenshots, but also dynamically adapting to changes as the interface evolves. We review existing works based on their usage of input features:
\begin{itemize*} 
\item \textbf{Structured Text}: 
early iterations~\citep{li2024sheetcopilot,wang2023enabling,wu2024copilot} of GUI agents powered by LLMs are constrained by the LLMs' limitation of processing only textual input. Consequently, these agents rely on converting GUI pages into structured textual representations, such as HTML, accessibility trees, or Document Object Model (DOM).
For web pages, some agents use HTML data as input or leverage the DOM to analyze pages' layout.
% The DOM offers a hierarchical representation of elements. 
The DOM provides a tree-like structure that organizes elements hierarchically.
To reduce input noise, 
Agent-E~\citep{abuelsaad2024agent} utilizes a DOM distillation technique to achieve more effective screenshot representations. \citet{tao2023webwise} introduce WebWISE, which iteratively generates small programs based on observations from filtered DOM elements and performs tasks in a sequential manner.

\item\textbf{Visual Screenshot}: with advancements in computer vision and VLMs, agents are now capable of leveraging visual data from screens to interpret their on-screen environments. A significant portion of research relies on Set-of-Mark (SoM)~\citep{yang2023set} prompting to improve the visual grounding capabilities. To enhance visual understanding, these methods frequently employ Optical Character Recognition (OCR) in conjunction with GUI element detection models, including ICONNet~\citep{sunkara2022towards} and DINO~\citep{liu2025grounding}. These algorithms are used to identify and delineate interactive elements through bounding boxes, which are subsequently mapped to specific image regions, enriching the agents' contextual comprehension. Some studies also improve the semantic grounding ability and understanding of elements by adding descriptions of these interactive elements in the screenshots. For example, SeeAct~\citep{zheng2024gpt} enhances fine-grained screenshot content understanding by associating visual elements with the content they represent in HTML web.

\item \textbf{Comprehensive Interface Modeling}: recently, certain works have employed structured text, visual snapshots, and semantic outlines of elements to attain a holistic understanding of external perception. For instance, \citet{gou2024navigating}
synthesize large-scale GUI element data and train a visual grounding model UGround to gain the associated references of elements in GUI pages on various platforms.
Similarly, OSCAR~\citep{wang2024oscar} utilizes an A11y tree generated by the Windows API for representing GUI components, incorporating descriptive labels to facilitate semantic grounding. Meanwhile, DUAL-VCR~\citep{kil2024dual} captures both the visual features of the screenshot and the descriptions of associated HTML elements to obtain a robust representation of the visual screenshot.
\end{itemize*}

Another important point is the ability to interact in real-time. GUIs are inherently dynamic, with elements frequently changing in response to user actions or system processes. GUI agents must continuously monitor these changes to maintain an up-to-date understanding of the interface's state. This real-time perception is critical for ensuring that agents can respond promptly and accurately to evolving conditions. For instance, if a loading spinner appears, the agent should recognize it as an indication of a pending process and adjust its actions accordingly. Similarly, agents must detect and handle scenarios where the interface becomes unresponsive or behaves unexpectedly.

By effectively combining these above aspects, a robust perception system ensures that the GUI agent can maintain situational awareness and respond appropriately to the evolving state of the user interface, aligning its actions with the user's goals and the application's requirements. However, privacy concerns and the additional perceptual noise introduced by the DOM make it challenging to extend pure text descriptions and hybrid text-visual perceptions to any GUI environment. Hence, similar to human interaction with their surroundings, a native agent model should directly comprehend the external environment through visual perception and ground their actions to the original screenshot accurately. By doing so, the native agent model can generalize various tasks and improve the accuracy of actions at each step.

\paragraph{Action}
Effective action mechanisms must be versatile, precise, and adaptable to various GUI contexts. Key aspects include:

\begin{itemize*} \item \textbf{Unified and Diverse Action Space:}
GUI agents~\citep{gur2023real,bonatti2024windowsagentarenaevaluating} operate across multiple platforms, including mobile devices, desktop applications, and web interfaces, each with distinct interaction paradigms. Establishing a unified action space abstracts platform-specific actions into a common set of operations such as \texttt{click}, \texttt{type}, \texttt{scroll}, and \texttt{drag}. Additionally, integrating actions from language agents—such as API calls~\citep{chen2024octopus,li2024sheetcopilot,li2023zero}, code interpretation~\citep{wu2024copilot}, and Command-Line Interface (CLI)~\citep{mei2024aios} operations—enhances agent versatility. Actions can be categorized into atomic actions, which execute single operations, and compositional actions, which sequence multiple atomic actions to streamline task execution. Balancing atomic and compositional actions optimizes efficiency and reduces cognitive load, enabling agents to handle both simple interactions and the coordinated execution of multiple steps seamlessly.

\item \textbf{Challenges in Grounding Coordinates:}  
accurately determining coordinates for actions like clicks, drags, and swipes is challenging due to variability in GUI layouts~\citep{he2024webvoyager,burger2020mobile}, differing aspect ratios across devices, and dynamic content changes. Different devices' aspect ratios can alter the spatial arrangement of interface elements, complicating precise localization. Grounding coordinates requires advanced techniques to interpret visual cues from screenshots or live interface streams accurately.
\end{itemize*}

Due to the similarity of actions across different operational spaces, agent models can standardize actions from various GUI contexts into a unified action space. Decomposing actions into atomic operations reduces learning complexity, facilitating faster adaptation and transfer of atomic actions across different platforms.

\paragraph{Reasoning with System 1\&2 Thinking}
Reasoning is a complex capability that integrates a variety of cognitive functions. Human interaction with GUIs relies on two distinct types of cognitive processes~\citep{groves1970habituation}: \textbf{system 1} and \textbf{system 2} thinking.

\begin{itemize*}
    \item \textbf{System 1} refers to fast, automatic, and intuitive thinking, typically employed for simple and routine tasks, such as clicking a familiar button or dragging a file to a folder without conscious deliberation.
    \item \textbf{System 2} encompasses slow, deliberate, and analytical thinking, which is crucial for solving complex tasks, such as planning an overall workflow or reflecting to troubleshoot errors.
\end{itemize*}

Similarly, autonomous GUI agents must develop the ability to emulate both system 1 and system 2 thinking to perform effectively across a diverse range of tasks. By learning to identify when to apply rapid, heuristic-based responses and when to engage in detailed, step-by-step reasoning, these agents can achieve greater efficiency, adaptability, and reliability in dynamic environments.

\textbf{System 1 Reasoning} represents the agent's ability to execute fast, intuitive responses by identifying patterns in the interface and applying pre-learned knowledge to observed situations. 
This form of reasoning mirrors human interaction with familiar elements of a GUI, such as recognizing that pressing “Enter” in a text field submits a form or understanding that clicking a certain button progresses to the next step in a workflow. 
These heuristic-based actions enable agents to respond swiftly and maintain operational efficiency in routine scenarios. However, the reliance on pre-defined mappings limits the scope of their decision-making to immediate, reactive behaviors. 
For instance, models such as large action models~\citep{wu2024atlas,wang2024mobile} excel at generating quick responses by leveraging environmental observations, but they often lack the capacity for more sophisticated reasoning. This constraint becomes particularly evident in tasks requiring the planning and execution of multi-step operations, which go beyond the reactive, one-step reasoning of system 1. Thus, while system 1 provides a foundation for fast and efficient operation, it underscores the need for agents to evolve toward more deliberate and reflective capabilities seen in system 2 reasoning.

\textbf{System 2 Reasoning} represents deliberate, structured, and analytical thinking, enabling agents to handle complex, multi-step tasks that go beyond the reactive behaviors of system 1. Unlike heuristic-based reasoning, system 2 involves explicitly generating intermediate thinking processes, often using techniques like Chain-of-Thought (CoT)~\citep{wei2022chain} or ReAct~\citep{yao2022react}, which bridge the gap between simple actions and intricate workflows. This paradigm of reasoning is composed of several essential components.

\begin{itemize*}
    \item First, \textbf{task decomposition} focuses on formulating plannings to achieve overarching objectives by decomposing tasks into smaller, manageable sub-tasks~\citep{dagan2023dynamicplanningllm, Song_2023_ICCV, huang2024understandingplanningllmagents}. For example, completing a multi-field form involves a sequence of steps like entering a name, address, and other details, all guided by a well-structured plan.

    \item Second, \textbf{long-term consistency} is critical during the entire task completion process. By consistently referring back to the initial objective, agent models can effectively avoid any potential deviations that may occur during complex, multi-stage tasks, thus ensuring coherence and continuity from start to finish.
    
    \item Third, \textbf{milestone recognition} 
    allows the agent model to estimate the current state of progress, analyze observations, and determine the subsequent goals. This ensures that multi-step workflows are executed effectively without losing direction.

    \item Fourth, \textbf{trial and error} endows agent models with additional opportunities to hypothesize, test, and assess potential actions, thereby enhancing the precision of decision-making, particularly in ambiguous and complex scenarios.

    \item Finally, \textbf{reflection} equips agent models with the capability to evaluate past actions, identify mistakes, and make adjustments to improve future performance~\citep{shinn2023reflexion, renze2024selfreflectionllmagentseffects}. This iterative process enhances reliability and helps prevent repetitive errors.
\end{itemize*}
The development of \ourmodel places a strong emphasis on equipping the model with robust system 2 reasoning capabilities, allowing it to address complex tasks with greater precision and adaptability. By integrating high-level planning mechanisms, \ourmodel excels at decomposing overarching goals into smaller, manageable sub-tasks. This structured approach enables the model to systematically handle intricate workflows that require coordination across multiple steps. Additionally, \ourmodel incorporates a long-form CoT reasoning process, which facilitates detailed intermediate thinking before executing specific actions. Furthermore, \ourmodel adopts reflection-driven training process. By incorporating reflective thinking, the model continuously evaluates its past actions, identifies potential mistakes, and adjusts its behavior to improve performance over time. The model's iterative learning method yields significant benefits, enhancing its reliability and equipping it to navigate dynamic environments and unexpected obstacles.

\paragraph{Memory}
The memory is mainly used to store the supported explicit knowledge and historical  experience that the agent refers to when making decisions. 
For agent frameworks, an additional memory module is often introduced to store previous interactions and task-level knowledge. Agents then retrieve and update these memory modules during decision-making progress. The memory module can be divided into two categories:
\begin{itemize*}
    \item \textbf{Short-term Memory:} this serves as a temporary repository for task-specific information, capturing the agent's immediate context. 
    This includes the agent's action history, current state details, and the ongoing execution trajectory of the task, enabling real-time situational awareness and adaptability. 
    By semantically processing contextual screenshots, CoAT~\citep{zhang2024android} extracts key interface details, thereby enhancing comprehension of the task environment. CoCo-Agent~\citep{ma2024comprehensive} records layouts and dynamic states through Comprehensive Environment Perception (CEP). 
    \item \textbf{Long-term Memory:} 
    it operates as a long-term data reserve, capturing and safeguarding records of previous interaction, tasks, and background knowledge. It retains details such as execution paths from prior tasks, offering a comprehensive knowledge base that supports reasoning and decision-making for future tasks. By integrating accumulated knowledge that contains user preferences and task operation experiences, OS-copilot~\citep{wu2024copilot} refines its task execution over time to better align with user needs and improve overall efficiency. Cradle~\citep{tan2024towards} focuses on enhancing the multitasking abilities of foundational agents by equipping them with the capability to store and utilize task execution experiences. \citet{song2024beyond} introduce a framework for API-driven web agents that leverage task-specific background knowledge to perform complex web operations. 
    %By using APIs to access and retrieve web-based knowledge, the system Infogent~\citep{reddy2024infogent} empowers agents to utilize online experiences and contextual information for tasks efficiently.
\end{itemize*}

Memory reflects the capability to leverage background knowledge and input context.
The synergy between short-term and long-term memory storage significantly enhances the efficiency of an agent's decision-making process.
Native agent models, unlike agent frameworks, encode long-term operational experience of tasks within their internal parameters, converting the observable interaction process into implicit, parameterized storage. Techniques such as In-Context Learning (ICL) or CoT reasoning can be employed to activate this internal memory.

\subsection{Capability Evaluation}
\label{sec:cap_evaluation}
% In this section, we present the evaluation paradigm of GUI agents.
% To assess the performance of GUI agents, 
To evaluate the effectiveness of GUI agents,
numerous benchmarks have been meticulously designed, focusing on various aspects of capabilities such as perception, grounding, and agent capabilities. Specifically, \textbf{Perception Evaluation} reflects the degree of understanding of GUI knowledge. \textbf{Grounding Evaluation} verifies whether agents can accurately locate coordinates in diverse GUI layouts. Agent capabilities can be primarily divided into two categories: \textbf{Offline Agent Capability Evaluation}, which is conducted in a predefined and static environment and mainly focuses on assessing the individual steps performed by GUI agents, and \textbf{Online Agent Capability Evaluation}, which is performed in an interactive and dynamic environment and evaluates the agent's overall capability to successfully complete the task.

\paragraph{Perception Evaluation}
Perception evaluation assesses agents' understanding of user interface (UI) knowledge and their awareness of the environment. For instance, VisualWebBench~\citep{liu2024visualwebbench} focuses on agents' web understanding capabilities, while WebSRC~\citep{chen2021websrc} and ScreenQA~\citep{hsiao2022screenqa} evaluate web structure comprehension and mobile screen content understanding through question-answering (QA) tasks. Additionally, GUI-World~\citep{chen2024guiworlddatasetguiorientedmultimodal} offers a wide range of queries in multiple-choice, free-form, and conversational formats to assess GUI understanding.
Depending on the varying question formats, a range of metrics are employed. For instance, accuracy is utilized for multiple-choice question (MCQ) tasks as the key metric, and in the case of captioning or Optical Character Recognition (OCR) tasks, the ROUGE-L metric is adopted to evaluate performance.

\paragraph{Grounding Evaluation}
Given an instructions, grounding evaluation focuses on the ability to precisely locate GUI elements. ScreenSpot~\citep{cheng2024seeclickharnessingguigrounding} evaluates single-step GUI grounding performance across multiple platforms. ScreenSpot v2~\citep{wu2024atlas}, a re-annotated version, addresses annotation errors present in the original ScreenSpot. ScreenSpot Pro~\citep{li2024screenspot-pro} facilitates grounding evaluation by incorporating real-world tasks gathered from diverse high-resolution professional desktop environments.
Metrics for grounding evaluation are usually determined based on whether the model's predicted location accurately lies within the bounding box of the target element.
% Metrics for grounding evaluation are typically defined as the proportion of test samples in which the model's predicted location accurately falls within the bounding box of the ground truth element.

\paragraph{Offline Agent Capability Evaluation}
Offline evaluation measures the performance of GUI agents in static, pre-defined environments. Each environment typically includes an input instruction and the current state of the environment (e.g., a screenshot or a history of previous actions), requiring agents to produce the correct outputs or actions. These environments remain consistent throughout the evaluation process.
Numerous offline evaluation benchmarks, including 
AITW~\citep{10.5555/3666122.3668731},
Mind2Web~\citep{deng2023mind2webgeneralistagentweb}, 
MT-Mind2Web~\citep{deng2024multiturninstructionfollowingconversational}, AITZ~\citep{zhang2024androidzoochainofactionthoughtgui}, AndroidControl~\citep{li2024on}, and GUI-Odyssey~\citep{lu2024gui}, provide agents with a task description, a current screenshot, and previous actions history, aimed at enabling accurate prediction of the next action.
% with the primary objective to accurately predict the next action.
These benchmarks commonly employ step-level metrics 
% to evaluate the step-wise performance of GUI agents
, providing fine-grained supervision of their specific behaviors. For instance, the Action-Matching Score~\citep{10.5555/3666122.3668731,zhang2024androidzoochainofactionthoughtgui, li2024on, lu2024gui}  considers an action correct solely when both the type of action and its specific details (e.g. arguments like typed content or scroll direction) are consistent with the ground truth.
% deems an action correct only if both the action type and action details (e.g., arguments such as scroll direction, typed text, clicked position, and pressed button) align with the ground truth.
% Some benchmarks~\citep{seq2act, 10.1007/978-3-031-20074-8_18}
% , such as PIXELHELP~\citep{seq2act} and MoTIF~\citep{10.1007/978-3-031-20074-8_18},
Some benchmarks~\citep{seq2act, 10.1007/978-3-031-20074-8_18} demand that agents produce a series of automatically executable actions from provided instructions and screenshots. These benchmarks predominantly assess performance using task-level metrics, which determine task success by whether the output results precisely match the pre-defined labels, like the complete and partial action sequence matching accuracy~\citep{seq2act, 10.1007/978-3-031-20074-8_18, 10.5555/3666122.3668731}.

\paragraph{Online Agent Capability Evaluation}
Online evaluation facilitates dynamic environments, each designed as an interactive simulation that replicates real-world scenarios. In these environments, GUI agents can modify environmental states by executing actions in real time.
These dynamic environments span various platforms:
(1) Web: WebArena~\citep{zhou2024webarenarealisticwebenvironment} and MMInA~\citep{zhang2024mminabenchmarkingmultihopmultimodal} provide realistic web environments.
(2) Desktop: OSWorld~\citep{xie2024osworldbenchmarkingmultimodalagents}, OfficeBench~\citep{wang2024officebenchbenchmarkinglanguageagents}, ASSISTGUI~\citep{gao2024assistguitaskorienteddesktopgraphical}, and WindowsAgentArena~\citep{bonatti2024windowsagentarenaevaluating} operate within real computer desktop environments.
(3) Mobile: AndroidWorld~\citep{rawles2024androidworlddynamicbenchmarkingenvironment}, LlamaTouch~\citep{zhang2024llamatouch}, and B-MOCA~\citep{lee2024benchmarkingmobiledevicecontrol} are built on mobile operating systems such as Android.
To assess performance in online evaluation, task-level metrics are employed, providing a comprehensive measure of the agents' effectiveness. 
 Specifically, in the realm of online agent capability evaluation, these task-level metrics primarily determine task success based on whether an agent successfully reaches a goal state. This verification process checks whether the intended outcome achieved or if the resulting outputs precisely align with the labels~\citep{zhou2024webarenarealisticwebenvironment,xie2024osworldbenchmarkingmultimodalagents,wang2024officebenchbenchmarkinglanguageagents,gao2024assistguitaskorienteddesktopgraphical}.
\section{\ourmodel}
\label{sec:method}
In this section, we introduce \ourmodel, a native GUI agent model designed to operate without reliance on cumbersome manual rules or the cascaded modules typical of conventional agent frameworks. \ourmodel directly perceives the screenshot, applies reasoning processes, and generates valid actions autonomously. Moreover, \ourmodel can learn from prior experience, iteratively refining its performance by leveraging environment feedback.

In the following, we begin by describing the overall architecture of \ourmodel (\cref{sec:arch_overview}), followed by how we enhance its perception (\cref{sec:perception}) and action (\cref{sec:action}) capabilities. Then we concentrate on how to infuse system-2 reasoning capabilities into \ourmodel (\cref{sec:reasoning}) and iterative improvement through experience learning (\cref{sec:memory}).
\begin{figure}
    \centering
    \includegraphics[width=0.75\linewidth]{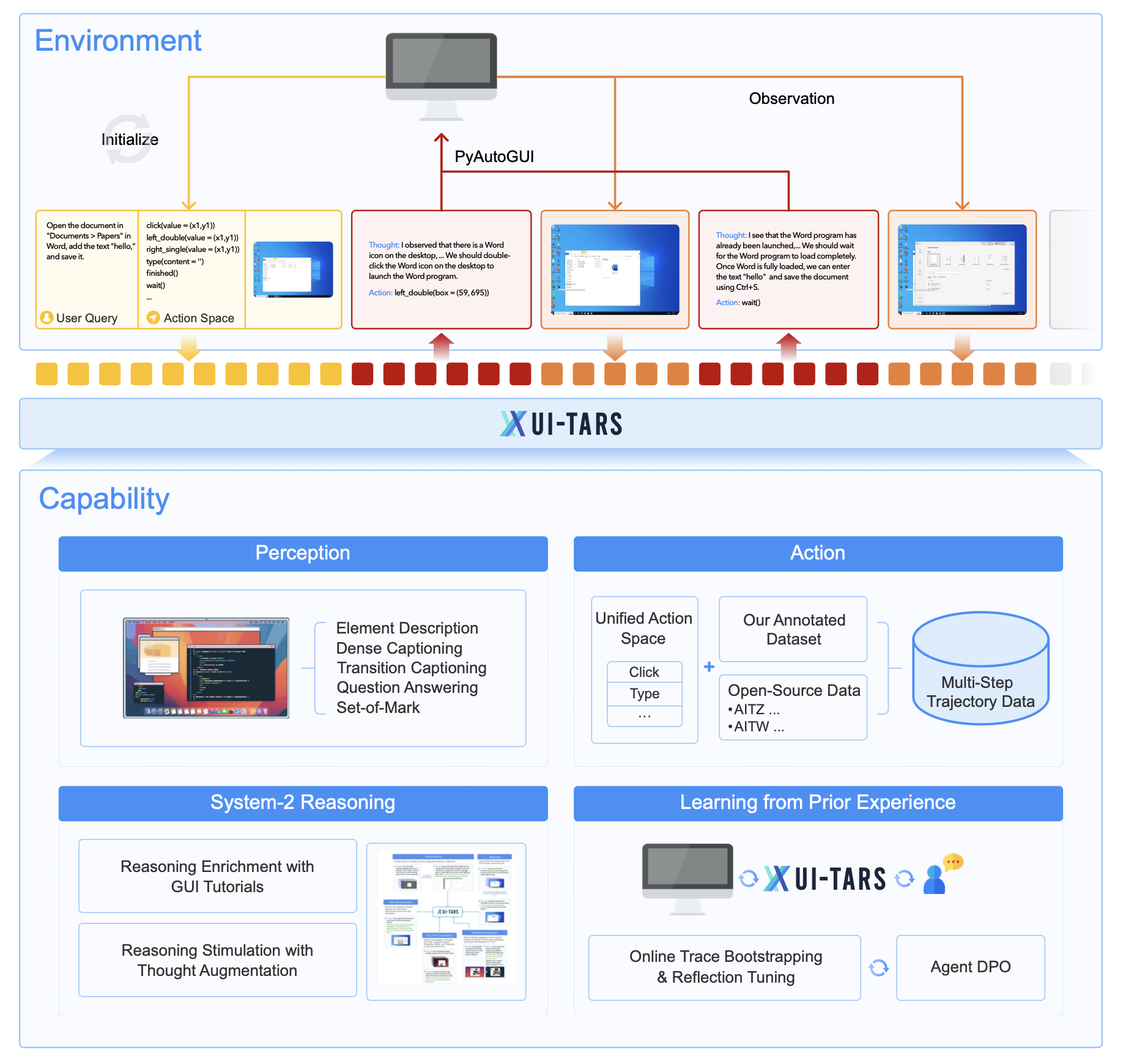}
    \caption{\small{Overview of UI-TARS. We illustrate the architecture of the model and its core capabilities.}}
    \label{model_arc}
\end{figure}

\subsection{Architecture Overview}
\label{sec:arch_overview}
As illustrated in Figure~\ref{model_arc}, given an initial task instruction, \ourmodel iteratively receives observations from the device and performs corresponding actions to accomplish the task. This sequential process can be formally expressed as:
\begin{equation}
(\text{instruction}, (o_1, a_1), (o_2, a_2), \cdots, (o_n, a_n)),
\end{equation}
where \(o_i\) denotes the observation (device screenshot) at time step \(i\), and \(a_i\) represents the action executed by the agent. At each time step, \ourmodel takes as input the task instruction, the history of prior interactions \((o_1, a_1, \cdots, o_{i-1}, a_{i-1})\), and the current observation \(o_i\). Based on this input, the model outputs an action \(a_i\) from the predefined action space. After executing the action, the device provides the subsequent observation, and these processes iteratively continue.

To further enhance the agent’s reasoning capabilities and foster more deliberate decision-making, we integrate a reasoning component in the form of ``thoughts'' \(t_i\), generated before each action \(a_i\). These thoughts reflect the reflective nature of “System 2” thinking. They act as a crucial intermediary step, guiding the agent to reconsider previous actions and observations before moving forward, thus ensuring that each decision is made with intentionality and careful consideration.

This approach is inspired by the ReAct framework~\citep{yao2022react}, which introduces a similar reflective mechanism but in a more straightforward manner. In contrast, our integration of “thoughts” involves a more structured, goal-oriented deliberation. These thoughts are a more explicit reasoning process that guides the agent toward better decision-making, especially in complex or ambiguous situations. The process can now be formalized as:
\begin{equation}
(\text{instruction}, (o_1, t_1, a_1), (o_2, t_2, a_2), \cdots, (o_n, t_n, a_n)),
\end{equation}
these intermediate thoughts guide the model’s decision-making and enable more nuanced and reflective interactions with the environment.

In order to optimize memory usage and maintain efficiency within the typically constrained token budget (e.g., 32k sequence length), we limit the input to the last $N$ observations. This constraint ensures the model remains capable of handling the necessary context without overwhelming its memory capacity. The full history of previous actions and thoughts is retained as short-term memory. \ourmodel predicts the thought \(t_n\) and action \(a_n\) outputs iteratively, conditioned on both the task instruction and the previous interactions:
\begin{equation}
P(t_n, a_n \mid \text{instruction}, t_1, a_1, \cdots, (o_{n-i}, t_{n-i}, a_{n-i})_{i=1}^N, o_n).
\label{eq:last_N}
\end{equation}

\subsection{Enhancing GUI Perception}
\label{sec:perception}

Improving GUI perception presents several unique challenges: (1) Screenshot Scarcity: while large-scale general scene images are widely available, GUI-specific screenshots are relatively sparse. (2) Information Density and Precision Requirement: GUI images are inherently more information-dense and structured than general scene images, often containing hundreds of elements arranged in complex layouts. Models must not only recognize individual elements but also understand their spatial relationships and functional interactions. Moreover, many elements in GUI images are small (e.g., 10$\times$10 pixel icons in a 1920$\times$1080 image), making it difficult to perceive and localize these elements accurately. Unlike traditional frameworks that rely on separate, modular perception models, native agents overcome these challenges by directly processing raw input from GUI screenshots. This approach enables them to scale better by leveraging large-scale, unified datasets, thereby addressing the unique challenges of GUI perception with greater efficiency.

\begin{figure}[!t]
    \centering
    \includegraphics[width=\linewidth]{figures/Single-step-data-new.pdf}
    \caption{Data example of perception and grounding data.}
    \label{fig:perception_data_annotation}
\end{figure}

\paragraph{Screenshot Collection}
To address data scarcity and ensure diverse coverage, we built a large-scale dataset comprising screenshots and metadata from websites, apps, and operating systems. Using specialized parsing tools, we automatically extracted rich metadata—such as element type, depth, bounding box, and text content for each element—while rendering the screenshots. Our approach combined automated crawling and human-assisted exploration to capture a wide range of content. We included primary interfaces as well as deeper, nested pages accessed through repeated interactions. All data was logged in a structured format—(screenshot, element box, element metadata)—to provide comprehensive coverage of diverse interface designs. 

We adopt a bottom-up data construction approach, starting from individual elements and progressing to holistic interface understanding. By focusing on small, localized parts of the GUI before integrating them into the broader context, this approach minimizes errors while balancing precision in recognizing components with the ability to interpret complex layouts. Based on the collected screenshot data, we curated five core task data (Figure~\ref{fig:perception_data_annotation}):

\paragraph{Element Description}
To enhance recognizing and understanding specific elements within a GUI, particularly tiny elements, we focus on creating detailed and structured descriptions for each element. Such descriptions are based on metadata extracted using parsing tools and further synthesized by a VLM, covering four aspects: (1) \textbf{Element Type} (e.g., windows control types): we classify elements (e.g., buttons, text fields, scrollbars) based on visual cues and system information; (2) \textbf{Visual Description}, which describes the element’s appearance, including its shape, color, text content, and style, derived directly from the image; (3) \textbf{Position Information}: we describe the spatial position of each element relative to others; (4) \textbf{Element Function}, which describes the element's intended functionality and possible ways of interactions. We train \ourmodel to enumerate all visible elements within a screenshot and generate their element descriptions, conditioned on the screenshot.

\paragraph{Dense Captioning}
We train \ourmodel to understand the entire interface while maintaining accuracy and minimizing hallucinations. The goal of dense captioning is to provide a comprehensive, detailed description of the GUI screenshot, capturing not only the elements themselves but also their spatial relationships and the overall layout of the interface. For each recorded element in the screenshot, we first obtain their element descriptions. For embedded images, which often lack detailed metadata, we also generate their descriptive captions. After that, we integrate all the image and element descriptions into a cohesive, highly detailed caption that preserves the structure of the GUI layout using a VLM. During training, \ourmodel is given only the image and tasked with outputting the corresponding dense caption.

\paragraph{State Transition Captioning}
While dense captioning provides a comprehensive description of a GUI interface, it does not capture state transitions, particularly the subtle effects of actions (e.g., a tiny button being pressed) on the interface. To address this limitation, we train the model to identify and describe the differences between two consecutive
 screenshots and determine whether an action, such as a mouse click or keyboard input, has occurred. We also incorporate screenshot pairs that correspond to non-interactive UI changes (e.g., animations, screen refreshes, or background updates). During training, \ourmodel is presented with a pair of images and tasked with predicting the specific visual changes (and possible reasons) of the two images. In this way, \ourmodel learns the subtle UI changes, including both user-initiated actions and non-interactive transitions. This capability is crucial for tasks requiring fine-grained interaction understanding and dynamic state perception.

\paragraph{Question Answering (QA)}
While dense captioning and element descriptions primarily focus on understanding the layout and elements of a GUI, QA offers a more dynamic and flexible approach to integrating these tasks with reasoning capabilities. We synthesize a diverse set of QA data that spans a broad range of tasks, including interface comprehension, image interpretation, element identification, and relational reasoning. This enhances \ourmodel's capacity to process queries that involve a higher degree of abstraction or reasoning.

\paragraph{Set-of-Mark (SoM)}
We also enhance the Set-of-Mark (SoM) prompting ability~\citep{yang2023set} of \ourmodel. We draw visually distinct markers for parsed elements on the GUI screenshot based on their spatial coordinates. These markers vary in attributes such as form, color, and size, providing clear, intuitive visual cues for the model to locate and identify specific elements. In this way, \ourmodel better associates visual markers with their corresponding elements. We integrate SoM annotations with tasks like dense captioning and QA. For example, the model might be trained to describe an element highlighted by a marker.

\subsection{Unified Action Modeling and Grounding}
\label{sec:action}
The de-facto approach for improving action capabilities involves training the model to mimic human behaviors in task execution, i.e., behavior cloning~\citep{bain1995framework}. While individual actions are discrete and isolated, real-world agent tasks inherently involve executing a sequence of actions, making it essential to train the model on multi-step trajectories. This approach allows the model to learn not only how to perform individual actions but also how to sequence them effectively (system-1 thinking).

\paragraph{Unified Action Space}
Similar to previous works, we design a common action space that standardizes semantically equivalent actions across devices (Table~\ref{tab:actionspace}), such as “click” on Windows versus “tap” on mobile, enabling knowledge transfer across platforms. Due to device-specific differences, we also introduce optional actions tailored to each platform. This ensures the model can handle the unique requirements of each device while maintaining consistency across scenarios. We also define two terminal actions: \texttt{Finished()}, indicating task completion, and \texttt{CallUser()}, invoked in cases requiring user intervention, such as login or authentication.

\paragraph{Action Trace Collection}
A significant challenge in training models for task execution lies in the limited availability of multi-step trajectory data, which has historically been under-recorded and sparse. To address this issue, we rely on two primary data sources: (1) \textbf{our annotated dataset}: we develop a specialized annotation tool to capture user actions across various software and websites within PC environments. The annotation process begins with the creation of initial task instructions, which are reviewed and refined by annotators to ensure clarity and alignment with the intended goals. Annotators then execute the tasks, ensuring that their actions fulfill the specified requirements. Each task undergoes rigorous quality filtering; and (2) \textbf{open-source data}: we also integrate multiple existing datasets (MM-Mind2Web~\citep{
zheng2024gpt4visiongeneralistwebagent}, GUIAct~\citep{ chen2024guicoursegeneralvisionlanguage}, AITW~\citep{10.5555/3666122.3668731}, 
AITZ~\citep{zhang2024android}, 
AndroidControl~\citep{li2024on},
GUI-Odyssey~\citep{lu2024gui},
AMEX~\citep{chai2024amexandroidmultiannotationexpo}) and standardize them into a unified action space format. This involves reconciling varying action representations into a consistent template, allowing for seamless integration with the annotated data. In Table~\ref{tab:gui_opensource_data}, we list the basic statistics of our action trace data.

\begin{table*}[!t]
    \centering
    \begin{minipage}{0.55\textwidth}
        \centering
        \resizebox{\textwidth}{!}{%
        \begin{tabular}{cll}
            \toprule
            \multicolumn{1}{c}{\textbf{Environment}} & \multicolumn{1}{l}{\textbf{Action}}                      & \multicolumn{1}{l}{\textbf{Definition}}                                                                 \\ \midrule
            \multirow{7}{*}{\textbf{Shared}}         & Click(x, y)                 & Clicks at coordinates (x, y).                                                \\
                                            & Drag(x1, y1, x2, y2)        & Drags from (x1, y1) to (x2, y2).                                           \\
                                            & Scroll(x, y, direction)     & Scrolls at (x, y) in the given direction.                                   \\
                                            & Type(content)               & Types the specified content.                                               \\
                                            & Wait()                      & Pauses for a brief moment.                                                 \\
                                            & Finished()                  & Marks the task as complete.                                                \\
                                            & CallUser()                  & Requests user intervention.                                               \\ \midrule
            \multirow{3}{*}{\textbf{Desktop}}        & Hotkey(key)                 & Presses the specified hotkey.                                              \\
                                            & LeftDouble(x, y)            & Double-clicks at (x, y).                                                   \\
                                            & RightSingle(x, y)          & Right-clicks at (x, y).                                                    \\ \midrule
            \multirow{4}{*}{\textbf{Mobile}}         & LongPress(x, y)             & Long presses at (x, y).                                                    \\
                                            & PressBack()                 & Presses the ``back'' button.                                      \\
                                            & PressHome()                 & Presses the ``home'' button.                                      \\
                                            & PressEnter()                & Presses the ``enter'' key.                                        \\ \bottomrule
            \end{tabular}
        }
        \caption{\small{Unified action space for different platforms.}}
        \label{tab:actionspace}
    \end{minipage}
    \hfill
    \begin{minipage}{0.43\textwidth}
        \centering
        \resizebox{\textwidth}{!}{
    
\begin{tabular}{ll|cc|cc}
\toprule
\multicolumn{2}{c|}{\multirow{2}{*}{\textbf{Data Type}}} & \multicolumn{2}{c|}{\textbf{Grounding}} & \multicolumn{2}{c}{\textbf{MultiStep}} \\
\cmidrule(l){3-4} \cmidrule(l){5-6}
& & \textbf{Ele.} & \textbf{Ele./Image} & \textbf{Trace} & \textbf{avg steps} \\
\midrule
\multirow{3}{*}{\textbf{Open Source}} & \textbf{Web} & $14.8$M & $6.7$ & $6.4$k & $7.1$ \\
& \textbf{Mobile} & $2.5$M & $4.6$ & $145$k & $9.6$ \\
& \textbf{Desktop} & $1.1$M & $6.2$ & 0 & 0 \\
\midrule
\multicolumn{2}{c|}{\textbf{Ours}} & * & $7.5$ & * & $14.9$ \\
\bottomrule
\end{tabular}
    }
        \caption{\small{Basic statistics for grounding and multi-step action trace data, comparing both our annotated dataset and open-source data across different platforms (web, mobile, and desktop). We report the number of elements (\textit{Ele.}) and the number of action traces (\textit{Trace}).}}
        \label{tab:gui_opensource_data}
    \end{minipage}
\vspace{-0.3cm}
\end{table*}

\paragraph{Improving Grounding Ability}
Grounding, the ability to accurately locate and interact with specific GUI elements, is critical for actions like clicking or dragging. Unlike multi-step action data, grounding data is easier to scale because it primarily relies on the visual and position properties of elements, which can be efficiently synthesized or extracted~\citep{hong2024cogagent,gou2024navigating,wu2024atlas}. We train \ourmodel to directly predict the coordinates of the elements it needs to interact with. This involves associating each element in a GUI with its spatial coordinates and metadata. 

As described in \cref{sec:perception}, we collected screenshots and extracted metadata, including element type, depth, bounding boxes, and text content, using specialized parsing tools. For elements recorded with bounding boxes, we calculated the average of the corners to derive a single point coordinate, representing the center of the bounding box. To construct training samples, each screenshot is paired with individual element descriptions derived from metadata. The model is tasked with outputting relative coordinates normalized to the dimensions of the screen, ensuring consistency across devices with varying resolutions. For example, given the description ``red button in the top-right corner labeled Submit'', the model predicts the normalized coordinates of that button. This direct mapping between descriptions and coordinates enhances the model's ability to understand and ground visual elements accurately.

To further augment our dataset, we integrated open-source data (Seeclick~\citep{cheng2024seeclickharnessingguigrounding}, GUIAct~\citep{chen2024guicoursegeneralvisionlanguage}, MultiUI~\citep{liu2024harnessingwebpageuistextrich}, Rico-SCA~\citep{seq2act}, WidgetCaption~\citep{li2020widgetcaptioninggeneratingnatural}, MUG~\citep{li2022muginteractivemultimodalgrounding}, Rico Icon~\citep{sunkara2022towards}, CLAY~\citep{li2022learningdenoiserawmobile}, UIBERT~\citep{Bai2021UIBertLG}, OmniACT~\citep{kapoor2024omniactdatasetbenchmarkenabling}, AutoGUI~\citep{anonymous2024autogui}, OS-ATLAS~\citep{wu2024atlas}) and standardized them into our unified action space format. We provide the basic statiscs of the grounding data for training in Table~\ref{tab:gui_opensource_data}. This combined dataset enables \ourmodel to achieve high-precision grounding, significantly improving its effectiveness in actions such as clicking and dragging.
\subsection{Infusing System-2 Reasoning}
\label{sec:reasoning}
Relying solely on system-1 intuitive decision-making is insufficient to handle complex scenarios and ever-changing environments. Therefore, we aim for \ourmodel to combine system-2 level reasoning, flexibly planning action steps by understanding the global structure of tasks.

\paragraph{Reasoning Enrichment with GUI Tutorials}
The first step focuses on reasoning enrichment, where we leverage publicly available tutorials that interweave text and images to demonstrate detailed user interactions across diverse software and web environments. These tutorials provide an ideal source for establishing foundational GUI knowledge while introducing logical reasoning patterns inherent to task execution. 

We selected MINT~\citep{awadalla2024mint} and OmniCorpus~\citep{li2024omnicorpus}, two widely recognized image-text interleaved pre-training datasets, as our initial data sources. However, these datasets contain substantial noise, with only a small fraction aligning with GUI tutorial criteria. To extract high-quality tutorial data, we implemented a multi-stage data collection and filtering pipeline: (1) \textbf{Coarse-Grained Filtering}: to isolate tutorial-like content, we trained a \textit{fastText} classifier~\citep{joulin2016fasttext} using a manually curated positive set of high-quality tutorials and random samples from MINT and OmniCorpus as the negative set. The trained classifier was then applied to perform an initial screening, filtering out irrelevant samples and generating a candidate dataset. (2) \textbf{Fine-Grained Filtering}: to further refine the candidate dataset, we employed an LLM to identify and remove false positives. This step ensured the remaining samples conformed to the characteristics of GUI tutorials. The coarse and fine filtering processes were iterated over multiple rounds to maximize the recall rate of high-quality GUI tutorials. (3) \textbf{Deduplication and Data Refinement}: the filtered dataset was further refined to address duplicates, advertisements, and residual noise. Deduplication was performed using URL-based and Locality-Sensitive Hashing (LSH) methods. Finally, we prompt an LLM to rephrase all the textual content in the tutorial, refining the content while eliminating irrelevant or low-quality ones.

Through this multi-stage process, we curated approximately $\mathbf{6}$M high-quality GUI tutorials. On average, each tutorial contains $510$ text tokens and $3.3$ images. This data not only enhances the model’s understanding of GUI operations but also lays a robust foundation for infusing reasoning capabilities.

\begin{figure}[!t]
    \centering
    \includegraphics[width=0.9\linewidth]{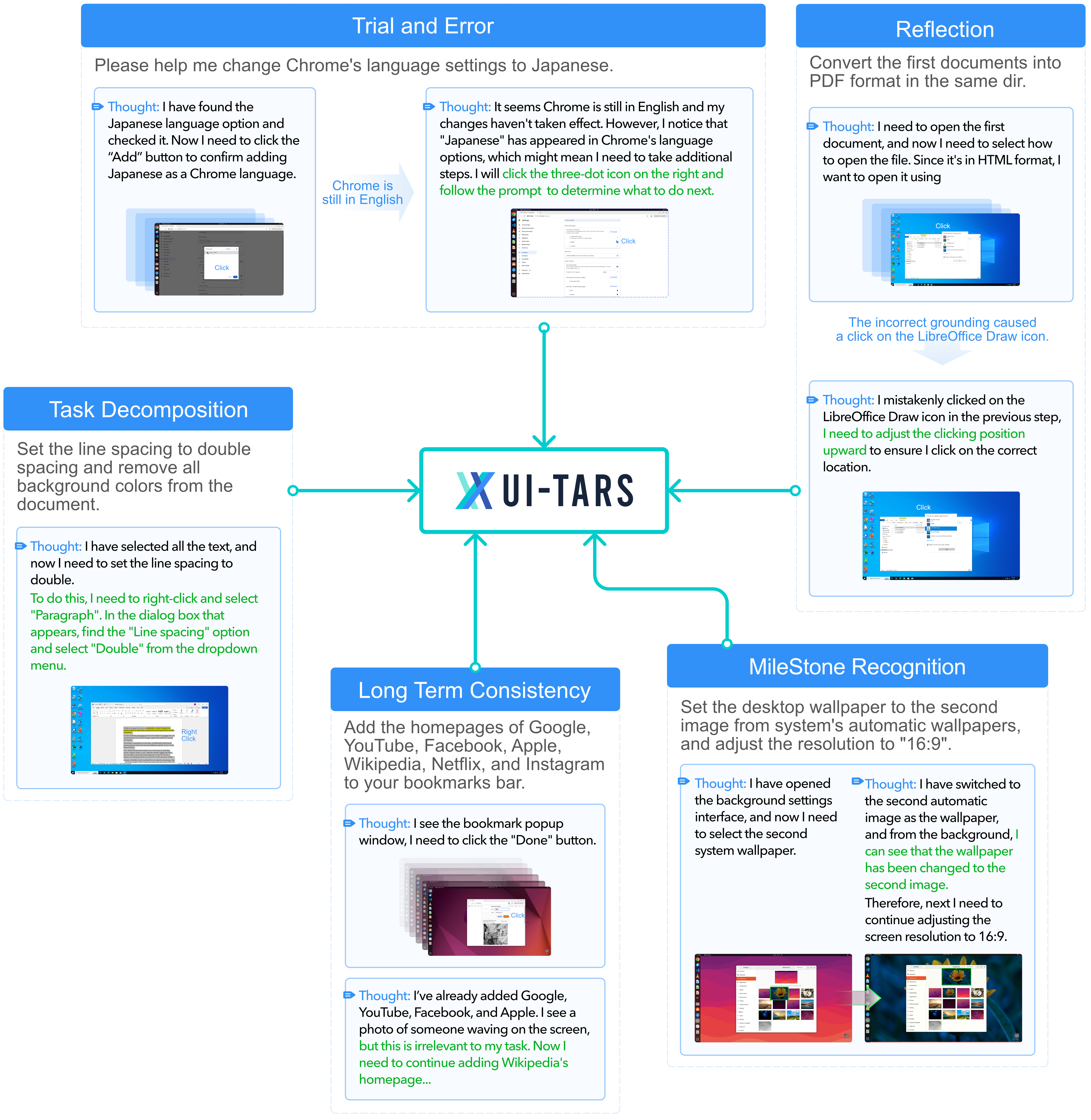}
    \caption{\small{Various reasoning patterns in our augmented thought.}}
    \label{fig:thought_annotation}
\end{figure}

\paragraph{Reasoning Stimulation with Thought Augmentation}

The action trace data we collect in \cref{sec:action} is inherently action-focused, containing sequences of observations and actions \((o_{i-1}, a_{i-1}, o_i, a_i, \dots)\) but lacking explicit reasoning thoughts. To stimulate reasoning capabilities of \ourmodel, we augment the dataset by annotating ``thoughts'' to bridge the gap between perception and action. This transforms the data format to \((o_{i-1}, t_{i-1}, a_{i-1}, o_{i}, t_i, a_i, \dots)\), where \(t\) represents the reasoning thought. These thoughts enable the model to express its decision-making process explicitly, fostering better alignment with task objectives. To construct these thoughts, we employ two annotation stages:

(1) \textbf{ActRe}~\citep{yang2024reactmeetsactrelanguage}: as shown in \eqref{equ:ActRe}, for each trace collected in \cref{sec:action}, we split them into multiple steps. For each step $n$, its thought \(t_n\) is generated iteratively by prompting a VLM with the previous context and the current target action \(a_n\). This method tries to make the generated thought logically grounded in the preceding context and aligned with the current action.

\begin{equation}
\left\{
\begin{array}{l}
{\color{green} t_n} = \text{VLM} (\text{instruction}, (o_1, t_1, a_1), (o_2, t_2, a_2), ..., o_n, {\color{red} a_n}) \\
t_{n+1} = \text{VLM} (\text{instruction}, (o_1, t_1, a_1), (o_2, t_2, a_2), ..., (o_n, {\color{green} t_n}, a_n), o_{n+1}, {\color{red} a_{n+1}}) \\
\vdots
\end{array}
\right.
\label{equ:ActRe}
\end{equation}

During ActRe annotation, we prompt the VLM toward exhibiting higher-order, system-2 reasoning, which involves deliberate, step-by-step decision-making and reflection. By promoting these reasoning patterns, we encourage the model to engage in thoughtful, long-term planning and reflection to solve complex tasks. As shown in Figure~\ref{fig:thought_annotation}, the reasoning patterns we prompt the VLM to follow include:
\begin{itemize*}
    \item \textbf{Task Decomposition}: guide the model to break complex tasks into smaller, manageable subtasks, enabling it to address intricate workflows step-by-step.
    \item \textbf{Long-term Consistency}: ensure the model maintains a consistent goal throughout a task, referencing the overall objective and operation history to avoid deviations during complex, multi-step tasks.
    \item \textbf{Milestone Recognition}: enable the model to recognize the completion of intermediate objectives, facilitating smooth transitions to subsequent goals.
    \item \textbf{Trial and Error}: equip the model to hypothesize, test, and evaluate potential actions, especially in ambiguous situations like verifying search results without directly interacting.
    \item \textbf{Reflection}: enable the model to identify and correct errors when operations fail, encouraging adaptability and error recovery through reflective reasoning.
\end{itemize*}
(2) \textbf{Thought Bootstrapping}: reverse annotation of thoughts conditioned on ground-truth actions (i.e., ActRe) can lead to false positives because the generated thoughts may appear to match the corresponding actions at a superficial level, without establishing a true causal relationship. Specifically, the reasoning process underlying the action may be overlooked, causing the thought to align with the action only by coincidence rather than through logical reasoning. This issue arises because the annotation process relies on knowing the action in advance, which may bias the thought to conform to the action rather than reflect the actual decision-making process leading to it.

To address this, we adopt a bootstrapping approach that generates thoughts without prior knowledge of the ground-truth action. By sampling multiple thought-action pairs, as shown in \eqref{equ:Bootstrapped_Thought}, we identify the thought that leads to the correct action, ensuring that the reasoning aligns causally with the chosen action. This approach produces higher-quality annotations because it forces the model to simulate a genuine decision-making process rather than merely justifying a pre-determined action ($\text{\ourmodel}_{\text{early}}$ means an early-stage model checkpoint).

\begin{equation}
\left\{
\begin{array}{l}
(\hat{t}_{n_i}, \hat{a}_{n_i})^\text{max-try}_{i=1} = \text{\ourmodel}_{\text{early}}(\text{instruction}, (o_1, t_1, a_1), (o_2, t_2, a_2), ..., o_n) \\
\text{Select} (\hat{t}_{n_i}, \hat{a}_{n_i}), \text{ where } \hat{a}_{n_i} = a_n
\end{array}
\right.
\label{equ:Bootstrapped_Thought}
\end{equation}

We annotate thoughts in both Chinese and English, expanding linguistic diversity. Although we augment thoughts for all traces, we also involve the vanilla action traces (without thought) during training.

\subsection{Learning from Prior Experience in Long-term Memory}
\label{sec:memory}

GUI agents face significant challenges in scaling to the level of LLMs, primarily due to the scarcity of large-scale, standardized, real-world \textit{process data} for GUI operations. While LLMs can leverage abundant textual data that captures diverse knowledge and reasoning patterns, process data detailing user interactions and decision-making sequences within GUI environments is rarely recorded or systematically organized. This lack of data impedes the ability of GUI agents to scale effectively and generalize across a wide range of tasks. One promising solution lies in learning from prior experiences stored in long-term memory. By capturing and retaining knowledge from previous tasks, agents can leverage these past experiences to inform their future decisions, making their actions more adaptive and efficient.

To facilitate this process, we enable \ourmodel to dynamically learn from interactions with real-world devices. Through semi-automated data collection, filtering, and refinement, the model continuously improves while minimizing the need for manual intervention. By leveraging long-term memory, \ourmodel builds on its accumulated knowledge, refining its performance over time and adapting to new tasks more efficiently. Each iteration of this process results in a more capable model.

\begin{figure}[!t]
    \centering
    \includegraphics[width=0.9\linewidth]{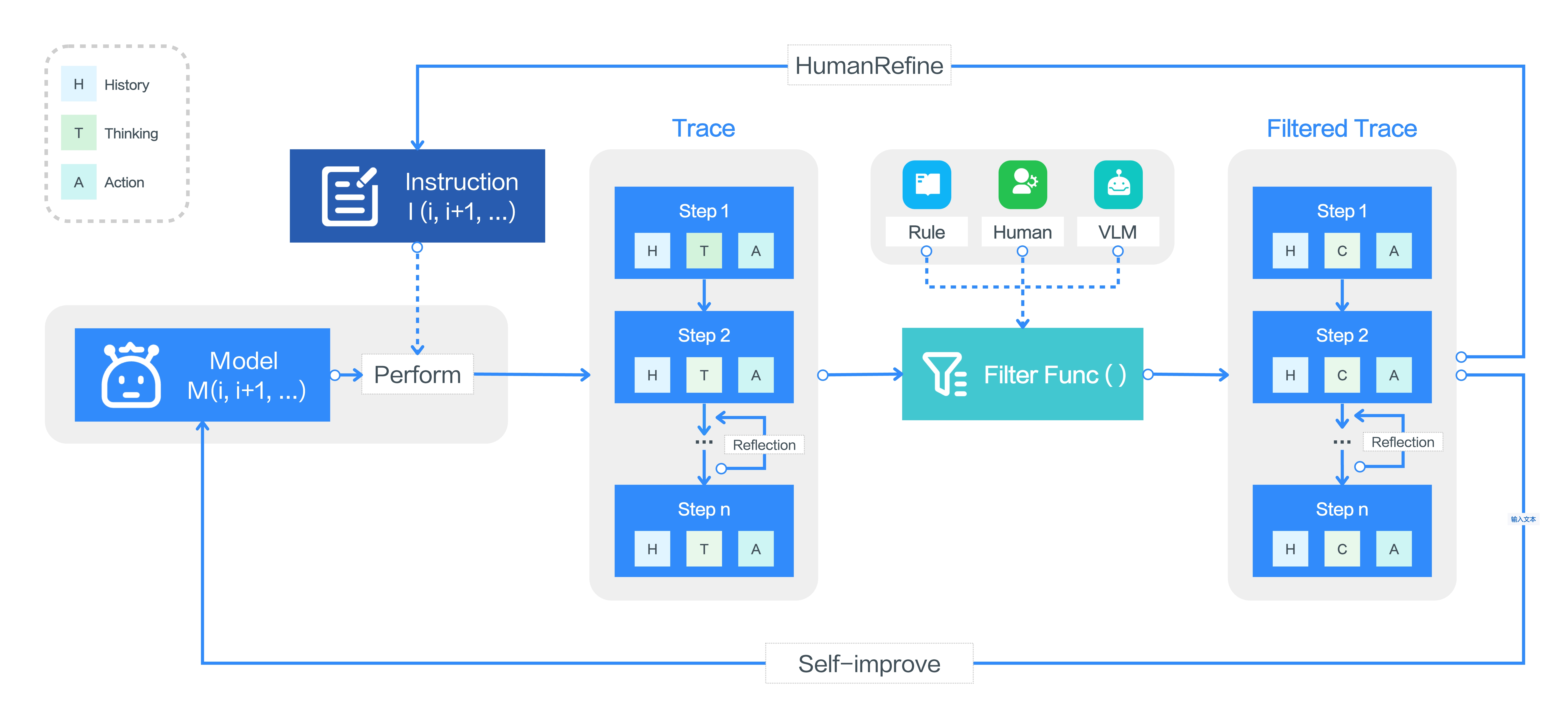}
    \caption{\small{Overview of the online bootstrapping process.}}
    \label{fig:online_bootstrap}
\end{figure}

\paragraph{Online Trace Bootstrapping}
As shown in Figure~\ref{fig:online_bootstrap}, we begin by obtaining a diverse set of task goals, combining both human-annotated and model-generated instructions. At iteration \(n\), the agent \(M_n\) executes these instructions $\mathcal{I}_n$ within the target GUI environments (e.g., a virtual PC), producing a raw set of traces:
\[
  \mathcal{T}_{\mathrm{raw},n} = \left\{ \left( o_1, t_1, a_1, o_2, t_2, a_2, \dots, o_n, t_n, a_n \right), \cdots \right\}.
\]
To ensure high-quality data, we apply a multi-level filtering function:
\[
  \mathrm{Filter}\left( \mathcal{T}_{\mathrm{raw},n}, \mathcal{I}_n \right) = \mathcal{T}_{\mathrm{filtered},n},
\]
which discards noisy or invalid traces through the following steps: 
(1) \textbf{Rule-Based Reward}: heuristic rules remove traces with obvious anomalies (e.g., redundant actions that do not alter the environment); 
(2) \textbf{VLM Scoring}: VLMs assign quality scores to the remaining traces, with traces scoring below a predefined threshold being removed;
(3) \textbf{Human Review}: part of the traces are further inspected by annotators, who identify the step where an error occurs, discard any subsequent actions, and retain only the valid prefix. \ourmodel leverages the resulting filtered trace set \(\mathcal{T}_{\mathrm{filtered},n}\) for self-improvement:
\[
  M_{n+1} = \mathrm{FineTune}\left( M_n, \mathcal{T}_{\mathrm{filtered},n} \right).
\]
For each round, we employ annotators to refine or expand the instruction set:
\[
  \mathcal{I}_{n+1} = \mathrm{HumanRefine}\left( \mathcal{I}_n, \mathcal{T}_{\mathrm{filtered},n} \right).
\]
We iterate the above process on hundreds of virtual PCs for multiple rounds, continuously leveraging the latest model \(M_{n+1}\) to generate new traces, thus expanding and refining the data.

\paragraph{Reflection Tuning} 
\label{para:reflection_annotation}
In realistic online deployments, agents often encounter situations where they get stuck due to a lack of self-reflection and error correction capabilities. For example, an agent might repeatedly click on an unresponsive button or attempt invalid operations due to misinterpretation of the interface. Without the ability to recognize these errors or adjust its strategy, the agent remains in a loop of ineffective actions, unable to progress toward the task objective. However, most offline datasets contain idealized, error-free trajectories because annotators ensure that each action meets expectations during the data labeling process. While such data helps reduce noise during model training, it also prevents the agent from learning how to recover from errors. To address this limitation, we propose a reflection tuning protocol that exposes the model to real-world errors made by itself with their corrections, enabling \ourmodel to learn how to recover from suboptimal decisions.

For an online trace generated by \ourmodel: $\mathcal{T} = \bigl(\text{instruction}, (o_1,t_1,a_1),(o_2, t_2,a_2),\dots,(o_t, t_t, a_t)\bigr)$, suppose that an error occurs at step \(\tau\), where the action \(a_\tau\) is deemed invalid or suboptimal. We asked annotators to identify this error and label the corrected thought and action \(t^*_\tau, a^*_\tau\). This results in an \textbf{error correction} trace pair:
\begin{equation}
\left\{
\begin{array}{l}
\mathcal{T}_{-} = \bigl(\text{instruction}, (o_1,t_1,a_1),(o_2, t_2,a_2),\dots,(o_\tau, {\color{red}t_\tau, a_\tau})\bigr) \\
\mathcal{T}_{+} = \bigl(\text{instruction}, (o_1,t_1,a_1),(o_2, t_2,a_2),\dots,(o_\tau, \color{green}t^*_\tau, \color{green}a^*_\tau)\bigr)
\end{array}
\right.
\label{equ:pair_data_1}
\end{equation}

Innovatively, we further require annotators to continue labeling the subsequent step based on the incorrect action \(a_\tau\), simulating a scenario where the error has already occurred. When determining the thought for the next step \(t^*_{\tau+1}\), annotators must acknowledge the impact of the previous mistake, compensate for its effects, and provide a correct action \(a^*_{\tau+1}\) to realign the task progress. For example, if the previous step intended to add a webpage to bookmarks but mistakenly clicked the close button, the next step should involve reopening the recently closed webpage to re-attempt clicking the bookmark button. Formally, we have a \textbf{post-reflection} trace pair:
\begin{equation}
\left\{
\begin{array}{l}
\mathcal{T}_{-} = \bigl(\text{instruction}, (o_1,t_1,a_1),(o_2, t_2,a_2),\dots,(o_\tau, {\color{red}t_\tau, a_\tau}), (o_{\tau+1}, t_{\tau+1}, a_{\tau+1})\bigr) \\
\mathcal{T}_{+} = \bigl(\text{instruction}, (o_1,t_1,a_1),(o_2, t_2,a_2),\dots,(o_\tau, {\color{red}t_\tau, a_\tau}),(o_{\tau+1}, \color{green}t^*_{\tau+1}, \color{green}a^*_{\tau+1})\bigr)
\end{array}
\right.
\label{equ:pair_data_2}
\end{equation}

We utilize the positive samples $\mathcal{T}_+$ for SFT training and calculate the loss only for the corrected steps (i.e., $(t^*_\tau, a^*_\tau)$ and $(t^*_{\tau+1}, a^*_{\tau+1})$), while the error steps (i.e., $(t_\tau, a_\tau)$) are not considered for training. Through this process, \ourmodel gradually improves its ability to recognize and recover from errors, enabling it to make effective adjustments when faced with imperfect or uncertain conditions. Cultivating this reflective ability enhances the agent's adaptability to dynamic environments and tasks.

\paragraph{Agent DPO}  
During online bootstrapping, a large number of erroneous steps (negative examples) are naturally generated. However, SFT only utilizes the corrected steps (i.e., ``positive'' examples), while ignoring the negative samples, which limits its ability to explicitly guide the agent away from suboptimal actions. To address this limitation, we turn to Direct Preference Optimization (DPO)~\citep{rafailov2024direct}, which leverages both the corrected and erroneous actions by introducing a reference-based objective. This approach optimizes \ourmodel by directly encoding a preference for corrected actions over erroneous ones, thereby making better use of the available data.

Consider a state \(s_\tau\) where the agent initially performed an incorrect action \(a_\tau\), which was later corrected to a preferred action \(a_\tau'\). Here, the state \(s_\tau\) consists of the instruction and its interaction history up to the current step \((o_1, t_1, a_1, \ldots, o_{\tau-1}, t_{\tau-1}, a_{\tau-1})\). This comprehensive representation provides the necessary context for the agent to make informed decisions. The key idea is to define a \textit{preference likelihood} that quantifies how much the model favors the corrected action \(a_\tau'\) over the original action \(a_\tau\). Formally, we define a learned reward function \(r_\theta(s, a)\) that estimates the desirability of taking action \(a\) in state \(s\). Based on Bradley-Terry model~\citep{bradley1952rank}, we can express the pairwise preference likelihood as:  
\[
P_\theta(a_\tau' \succ a_\tau \mid s_\tau)  
= \frac{\exp\bigl(r_\theta(s_\tau, a_\tau')\bigr)}  
{\exp\bigl(r_\theta(s_\tau, a_\tau)\bigr) + \exp\bigl(r_\theta(s_\tau, a_\tau')\bigr)},  
\]  
where \(a_\tau' \succ a_\tau\) indicates that \(a_\tau'\) is preferred over \(a_\tau\). The numerator represents the exponential of the reward assigned to the corrected action, while the denominator sums the exponentials of the rewards for both actions, ensuring that the likelihood is properly normalized.

DPO derives the analytical optimal policy given the reward function from the reinforcement learning (RL) objective with a KL-divergence constraint. We follow DPO to replace the reward function \(r_\theta\) with the optimal policy and directly optimize the DPO objective on the preference dataset:
\[
\mathcal{L}_{\mathrm{DPO}}(\theta)  
= - \mathop{\mathbb{E}}_\tau \left[ \log \sigma \left(\beta \log \frac{\pi_\theta (a_\tau' | s_\tau)}{\pi_{\mathrm{SFT}}(a_\tau' | s_\tau)} - \beta \log \frac{\pi_\theta (a_\tau | s_\tau)}{\pi_{\mathrm{SFT}}(a_\tau|s_\tau)} \right) \right]
\]

where \(\tau\) goes over all timesteps for which error-correction pairs are available. \(\pi_\theta\) denotes the optimal agent, \(\pi_\mathrm{SFT}\) denotes the SFT agent and \(\beta\) as a hyper-parameter controls the divergence between the optimal agent and the SFT agent. By minimizing the DPO loss, we fit the agent to increase the likelihood of the corrected actions and decrease the likelihood of the erroneous actions with an implicit reward function.

\subsection{Training}
To ensure a fair comparison with existing works such as Aguvis~\citep{xu2024aguvis} and OS-Atlas~\citep{wu2024atlas}, we use the same VLM backbone, Qwen-2-VL~\citep{wang2024qwen2vlenhancingvisionlanguagemodels}, and adopt a three-phase training process. This process refines the model’s capabilities across diverse GUI tasks, utilizing a total data size of approximately \textbf{50B} tokens. Each phase progressively incorporates higher-quality data to enhance the model’s performance on complex reasoning tasks.

\begin{itemize*}

\item \textbf{Continual Pre-training Phase}: we utilize the full set of data described in \cref{sec:method}, excluding the reflection tuning data, for continual pre-training with a constant learning rate. This foundational phase allows the model to learn all the necessary knowledge for automated GUI interaction, including perception, grounding, and action traces, ensuring robust coverage across diverse GUI elements and interactions. 

\item \textbf{Annealing Phase}: we then select high-quality subsets of perception, grounding, action trace, reflection tuning data for annealing. The annealing process gradually adjusts the model’s learning dynamics, promoting more focused learning and better optimization of its decision-making strategies in real-world GUI interaction scenarios. We denote the model trained after this phase as \ourmodel-SFT.

\item \textbf{DPO Phase}: finally, we employ annotated reflective pairs from online bootstrapping data for DPO training. During this process, the model refines its decision-making, reinforcing optimal actions while penalizing suboptimal ones. This process improves the model’s ability to make precise, context-aware decisions in real-world GUI interactions. The final model is denoted as \ourmodel-DPO.

\end{itemize*}

\newcommand{\VisualWebBench}{VisualWebBench}
\newcommand{\WebSRC}{WebSRC}
\newcommand{\SQAshort}{ScreenQA-short}
\newcommand{\ScreenSpot}{ScreenSpot}
\newcommand{\ScreenSpotVTwo}{ScreenSpot-V2}
\newcommand{\MMToWeb}{Multimodal-Mind2Web}
\newcommand{\AndroidControl}{Android Control}
\newcommand{\GUIOdyssey}{GUI Odyssey}
\newcommand{\OSWorld}{OSWorld}
\newcommand{\AndroidWorld}{AndroidWorld}
\newcommand{\ScreenSpotPro}{ScreenSpot-Pro}

\section{Experiment}
\label{sec:experiment}
In this section, we evaluate the performance of \ourmodel, trained on the dataset described in \cref{sec:method}, consisting of approximately \textbf{50B} tokens. We choose Qwen-2-VL~\citep{wang2024qwen2vlenhancingvisionlanguagemodels} as the base model for training and developed three model variants: \ourmodel-2B, \ourmodel-7B and \ourmodel-72B. Extensive experiments are conducted to validate the advantages of the proposed models. These experiments are designed to assess the models’ capabilities in three critical dimensions: perception, grounding, and agent capabilities.
Finally, we perform an ablation study to further investigate the impact of system 1 and system 2 reasoning on downstream tasks. We set the $N$ in Eq.~\ref{eq:last_N} to $5$ throughout this section. We evaluate both \ourmodel-SFT and \ourmodel-DPO for OSWorld in \cref{sec:exp_online}, as this benchmark benefits most from the iterative improvement from the DPO phase. For other benchmarks, however, we report the model trained after the annealing phase (i.e., \ourmodel-SFT).

\paragraph{Baseline}
 We compare \ourmodel with various baselines, including commercial models such as GPT-4o~\citep{hurst2024gpt}, Claude-3.5-Sonnet~\citep{claude-3-5-sonnet}, Gemini-1.5-Pro~\citep{team2024gemini}, and Gemini-2.0 (Project Mariner)~\citep{geimini}, as well as academic models from CogAgent~\citep{hong2024cogagent}, OminiParser~\citep{lu2024omniparser}, InternVL~\citep{chen2024internvl}, Aria-UI~\citep{yang2024aria}, Aguvis~\citep{xu2024aguvis}, OS-Atlas~\citep{wu2024atlas}, UGround~\citep{gou2024navigatingdigitalworldhumans}, ShowUI~\citep{lin2024showuivisionlanguageactionmodelgui}, SeeClick~\citep{cheng2024seeclickharnessingguigrounding}, the Qwen series models QwenVL-7B~\citep{bai2023qwenvlversatilevisionlanguagemodel}, Qwen2-VL (7B and 72B)~\citep{wang2024qwen2vlenhancingvisionlanguagemodels}, UIX-Qwen2-7B~\citep{liu2024harnessing} and Qwen-VL-Max~\citep{bai2023qwen}.

\subsection{Perception Capability Evaluation}
We evaluate the perception capabilities of the UI-TARS models using three key benchmarks: VisualWebBench~\citep{liu2024visualwebbench}, WebSRC~\citep{chen2021websrc}, and ScreenQA-short~\citep{hsiao2022screenqa}. VisualWebBench measures the model’s ability to understand and ground web elements, covering tasks like webpage QA, webpage OCR, and action prediction. UI-TARS models achieve outstanding results, with the 72B variant scoring $82.8$, significantly outperforming  closed-source models like GPT-4o ($78.5$) and Cluade 3.5 ($78.2$), as shown in Table \ref{tab:gui_exp_perception}.
For WebSRC and ScreenQA-short, which assess web structural comprehension and mobile screen content understanding through QA tasks, UI-TARS models show a clear advantage. WebSRC focuses on understanding the semantic content and layout of webpages in web contexts, while ScreenQA-short evaluates the interpretation of complex mobile screen layouts and interface-related questions. \ourmodel-7B achieves a leading score of $93.6$ on WebSRC, while \ourmodel-72B excells in ScreenQA-short with a score of $88.6$.
These results demonstrate the superior perception and comprehension capabilities of UI-TARS in web and mobile environments. Such perceptual ability lays the foundation for agent tasks, where accurate environmental understanding is crucial for task execution and decision-making.

\begin{table}[!t]
    \centering
    \small
    \resizebox{0.7\textwidth}{!}{%
    \begin{tabular}{lccc}
        \toprule
        \textbf{Model} & \textbf{\VisualWebBench} & \textbf{\WebSRC} & \textbf{\SQAshort} \\
        \midrule
        Qwen2-VL-7B~\citep{wang2024qwen2vlenhancingvisionlanguagemodels} & $73.3$ & $81.8$ & $84.9$ \\
        Qwen-VL-Max~\citep{bai2023qwenvlversatilevisionlanguagemodel} & $74.1$ & $91.1$ & $78.6$ \\
        Gemini-1.5-Pro~\citep{team2024gemini} & $75.4$ & $88.9$ & $82.2$ \\
        UIX-Qwen2-7B~\citep{wang2024qwen2} & $75.9$ & $82.9$ & $78.8$ \\
        Claude-3.5-Sonnet~\citep{claude-3-5-sonnet} & $78.2$ & $90.4$ & $83.1$ \\
        GPT-4o~\citep{hurst2024gpt} & $78.5$ & $87.7$ & $82.3$ \\
        \midrule
        \textbf{\ourmodel-2B} & $72.9$ & $89.2$ & $86.4$ \\
        \textbf{\ourmodel-7B} & $79.7$ & $\mathbf{93.6}$ & $87.7$ \\
        \textbf{\ourmodel-72B} & $\mathbf{82.8}$ & $89.3$ & $\mathbf{88.6}$ \\
        \bottomrule
    \end{tabular}
    }
    \caption{\small{Results on GUI Perception benchmarks.}}
    \label{tab:gui_exp_perception}
\end{table}

%     \end{minipage}
%     \hfill
%     \begin{minipage}{0.48\textwidth}
%         \caption{Results on GUI Grounding benchmarks. \textbf{Bold} text indicates the best-performing in each group respectively. \texttt{S.Spot} stands for \ScreenSpot.}
%         \centering
%         \resizebox{\textwidth}{!}{%
%         \begin{tabular}{r cccc}
%             \toprule
%             \multirow{2}{*}{\textbf{Model}} & \multicolumn{4}{c}{\textbf{GUI Grounding}} \\
%             \cmidrule(lr){2-5}
%             & \textbf{S.Spot} & \textbf{S.Spot-Pro} & \textbf{S.Spot-V2} & \textbf{RefExp} \\
%             \midrule
%             OS-Atlas-7B & $82.5$ & $18.9$ & $87.1$ & - \\
%             GPT-4o & $18.3$ & $0.8$ & - & - \\
%             Aria-UI & $82.4$ & $11.3$ & - & - \\
%             UGround-V1-7B & $86.3$ & $31.1$ & - & - \\
%             Qwen2-VL-7B & $69.2$ & $1.6$ & - & $18.8$ \\
%             UIX-Qwen2-7B & $55.2$ & - & - & $43.5$ \\
%             \midrule
%             \textbf{\ourmodel-7B} & $\mathbf{89.5}$ & $35.7$ & $\mathbf{91.6}$ & $83.7$ \\
%             \textbf{\ourmodel-72B} & $88.4$ & $\mathbf{38.1}$ & $90.3$ & $\mathbf{88.3}$ \\
%             \bottomrule
%         \end{tabular}
%         }
%         \label{tab:gui_exp_grounding}
%     \end{minipage}
% \vspace{-0.3cm}
% \end{table}

\begin{table}[!t]
    \centering
    \resizebox{\textwidth}{!}{%
    \begin{tabular}{lccccccccccccccccccccc}
        \toprule
        \multirow{2}{*}{\textbf{Agent Model}} & \multicolumn{3}{c}{\textbf{Development}} & \multicolumn{3}{c}{\textbf{Creative}} & \multicolumn{3}{c}{\textbf{CAD}} & \multicolumn{3}{c}{\textbf{Scientific}} & \multicolumn{3}{c}{\textbf{Office}} & \multicolumn{3}{c}{\textbf{OS}} & \multicolumn{3}{c}{\textbf{Avg}} \\
        \cmidrule(lr){2-4} \cmidrule(lr){5-7} \cmidrule(lr){8-10} \cmidrule(lr){11-13} \cmidrule(lr){14-16} \cmidrule(lr){17-19} \cmidrule(lr){20-22}
         & \textbf{Text} & \textbf{Icon} & \textbf{Avg} & \textbf{Text} & \textbf{Icon} & \textbf{Avg} & \textbf{Text} & \textbf{Icon} & \textbf{Avg} & \textbf{Text} & \textbf{Icon} & \textbf{Avg} & \textbf{Text} & \textbf{Icon} & \textbf{Avg} & \textbf{Text} & \textbf{Icon} & \textbf{Avg} & \textbf{Text} & \textbf{Icon} & \textbf{Avg} \\
        \midrule
QwenVL-7B~\citep{bai2023qwenvlversatilevisionlanguagemodel}        & $0.0$  & $0.0$  & $0.0$  & $0.0$  & $0.0$  & $0.0$  & $0.0$  & $0.0$  & $0.0$  & $0.7$  & $0.0$  & $0.4$  & $0.0$  & $0.0$  & $0.0$  & $0.0$  & $0.0$  & $0.0$  & $0.1$  & $0.0$  & \cellcolor{blue!10}$0.1$  \\
GPT-4o~\citep{hurst2024gpt}           & $1.3$  & $0.0$  & $0.7$  & $1.0$  & $0.0$  & $0.6$  & $2.0$  & $0.0$  & $1.5$  & $2.1$  & $0.0$  & $1.2$  & $1.1$  & $0.0$  & $0.9$  & $0.0$  & $0.0$  & $0.0$  & $1.3$  & $0.0$  & \cellcolor{blue!10}$0.8$  \\
SeeClick~\citep{cheng2024seeclickharnessingguigrounding}         & $0.6$  & $0.0$  & $0.3$  & $1.0$  & $0.0$  & $0.6$  & $2.5$  & $0.0$  & $1.9$  & $3.5$  & $0.0$  & $2.0$  & $1.1$  & $0.0$  & $0.9$  & $2.8$  & $0.0$  & $1.5$  & $1.8$  & $0.0$  & \cellcolor{blue!10}$1.1$  \\
Qwen2-VL-7B~\citep{wang2024qwen2vlenhancingvisionlanguagemodels}      & $2.6$  & $0.0$  & $1.3$  & $1.5$  & $0.0$  & $0.9$  & $0.5$  & $0.0$  & $0.4$  & $6.3$  & $0.0$  & $3.5$  & $3.4$  & $1.9$  & $3.0$  & $0.9$  & $0.0$  & $0.5$  & $2.5$  & $0.2$  & \cellcolor{blue!10}$1.6$  \\
% MiniCPM-V-7B     & $7.1$  & $0.0$  & $3.7$  & $2.0$  & $0.0$  & $1.2$  & $4.1$  & $1.6$  & $3.4$  & $8.3$  & $0.0$  & $4.7$  & $2.8$  & $3.8$  & $3.0$  & $3.7$  & $1.1$  & $2.6$  & $4.5$  & $0.7$  & \cellcolor{blue!10}$3.0$  \\
OS-Atlas-4B~\citep{wu2024atlas}      & $7.1$  & $0.0$  & $3.7$  & $3.0$  & $1.4$  & $2.3$  & $2.0$  & $0.0$  & $1.5$  & $9.0$  & $5.5$  & $7.5$  & $5.1$  & $3.8$  & $4.8$  & $5.6$  & $0.0$  & $3.1$  & $5.0$  & $1.7$  & \cellcolor{blue!10}$3.7$  \\
ShowUI-2B~\citep{qinghong2024showui}        & $16.9$ & $1.4$  & $9.4$  & $9.1$  & $0.0$  & $5.3$  & $2.5$  & $0.0$  & $1.9$  & $13.2$ & $7.3$  & $10.6$ & $15.3$ & $7.5$  & $13.5$ & $10.3$ & $2.2$  & $6.6$  & $10.8$ & $2.6$  & \cellcolor{blue!10}$7.7$  \\
CogAgent-18B~\citep{hong2024cogagent}     & $14.9$ & $0.7$  & $8.0$  & $9.6$  & $0.0$  & $5.6$  & $7.1$  & $3.1$  & $6.1$  & $22.2$ & $1.8$  & $13.4$ & $13.0$ & $0.0$  & $10.0$ & $5.6$  & $0.0$  & $3.1$  & $12.0$ & $0.8$  & \cellcolor{blue!10}$7.7$  \\
Aria-UI~\citep{yang2024aria}          & $16.2$ & $0.0$  & $8.4$  & $23.7$ & $2.1$  & $14.7$ & $7.6$  & $1.6$  & $6.1$  & $27.1$ & $6.4$  & $18.1$ & $20.3$ & $1.9$  & $16.1$ & $4.7$  & $0.0$  & $2.6$  & $17.1$ & $2.0$  & \cellcolor{blue!10}$11.3$ \\
UGround-7B~\citep{gou2024navigating}       & $26.6$ & $2.1$  & $14.7$ & $27.3$ & $2.8$  & $17.0$ & $14.2$ & $1.6$  & $11.1$ & $31.9$ & $2.7$  & $19.3$ & $31.6$ & $11.3$ & $27.0$ & $17.8$ & $0.0$  & $9.7$  & $25.0$ & $2.8$  & \cellcolor{blue!10}$16.5$  \\
Claude Computer Use~\citep{claude-computer-use}       & $22.0$ & $3.9$  & $12.6$ & $25.9$ & $3.4$  & $16.8$ & $14.5$ & $3.7$  & $11.9$ & $33.9$ & $15.8$  & $25.8$ & $30.1$ & $16.3$ & $26.9$ & $11.0$ & $4.5$  & $8.1$  & $23.4$ & $7.1$  & \cellcolor{blue!10}$17.1$  \\
OS-Atlas-7B~\citep{wu2024atlas}      & $33.1$ & $1.4$  & $17.7$ & $28.8$ & $2.8$  & $17.9$ & $12.2$ & $4.7$  & $10.3$ & $37.5$ & $7.3$  & $24.4$ & $33.9$ & $5.7$  & $27.4$ & $27.1$ & $4.5$  & $16.8$ & $28.1$ & $4.0$  & \cellcolor{blue!10}$18.9$ \\
UGround-V1-7B~\citep{gou2024navigating}      & - & -  & $35.5$ & - & -  & $27.8$ & - & -  & $13.5$ & - & -  & $38.8$ & - & -  & $48.8$ & - & -  & $26.1$ & - & -  & \cellcolor{blue!10}$31.1$ \\
\midrule
\textbf{\ourmodel-2B}    & $47.4$ & $4.1$ & $26.4$ & $42.9$ & $6.3$ & $27.6$ & $17.8$ & $4.7$ & $14.6$ & $56.9$ & $17.3$ & $39.8$ & $50.3$ & $17.0$ & $42.6$ & $21.5$ & $5.6$ & $14.3$ & $39.6$ & $8.4$ & \cellcolor{blue!10}$27.7$ \\
\textbf{\ourmodel-7B}    & $58.4$ & $12.4$ & $36.1$ & $50.0$ & $9.1$ & $32.8$ & $\mathbf{20.8}$ & $9.4$ & $\mathbf{18.0}$ & $63.9$ & $\mathbf{31.8}$ & $\mathbf{50.0}$ & $\mathbf{63.3}$ & $20.8$ & $53.5$ & $30.8$ & $\mathbf{16.9}$ & $24.5$ & $47.8$ & $16.2$ & \cellcolor{blue!10}$35.7$ \\
\textbf{\ourmodel-72B}    & $\mathbf{63.0}$ & $\mathbf{17.3}$ & $\mathbf{40.8}$ & $\mathbf{57.1}$ & $\mathbf{15.4}$ & $\mathbf{39.6}$ & $18.8$ & $\mathbf{12.5}$ & $17.2$ & $\mathbf{64.6}$ & $20.9$ & $45.7$ & $\mathbf{63.3}$ & $\mathbf{26.4}$ & $\mathbf{54.8}$ & $\mathbf{42.1}$ & $15.7$ & $\mathbf{30.1}$ & $\mathbf{50.9}$ & $\mathbf{17.5}$ & \cellcolor{blue!10}$\mathbf{38.1}$ \\
        \bottomrule
    \end{tabular}
    }
    \caption{\small{Comparison of various models on \ScreenSpotPro.}}
    \label{tab:screenspot_pro_comparison}
\end{table}

\subsection{Grounding Capability Evaluation}
To evaluate the grounding capabilities of the \ourmodel, we focus on three benchmarks: ScreenSpot Pro~\citep{li2024screenspot-pro}, ScreenSpot~\citep{cheng2024seeclickharnessingguigrounding}, and ScreenSpot v2~\citep{wu2024atlas}. These benchmarks assess the ability to understand and localize elements in GUIs. ScreenSpot Pro is designed for high-resolution professional environments, this benchmark includes expert-annotated tasks across 23 applications in five industries and three operating systems. It provides a rigorous assessment of model grounding performance in specialized, high-complexity scenarios. ScreenSpot and ScreenSpot v2 test GUI grounding across mobile, desktop, and web platforms. ScreenSpot evaluates models using both direct instructions and self-generated plans, while ScreenSpot v2 enhances the evaluation accuracy by correcting annotation errors.

\begin{table}[!t]
    \centering
     \resizebox{0.85\textwidth}{!}{%
    \begin{tabular}{llccccccccc}
        \toprule
        \multirow{2}{*}{\textbf{Method}} &  & \multicolumn{2}{c}{\textbf{Mobile}} & \multicolumn{2}{c}{\textbf{Desktop}} & \multicolumn{2}{c}{\textbf{Web}} & \multirow{2}{*}{\textbf{Avg}} \\
        \cmidrule(lr){3-4} \cmidrule(lr){5-6} \cmidrule(lr){7-8} % Partial horizontal line
        & & \textbf{Text} & \textbf{Icon/Widget} & \textbf{Text} & \textbf{Icon/Widget} & \textbf{Text} & \textbf{Icon/Widget} & \\
        \midrule
        \multicolumn{9}{l}{\textbf{Agent Framework}} \\
    \midrule
        \multirow{3}{*}{GPT-4~\citep{openai2023gpt4}} 
        & SeeClick~\citep{cheng2024seeclickharnessingguigrounding} & $76.6$ & $55.5$ & $68.0$ & $28.6$ & $40.9$ & $23.3$ & \cellcolor{blue!10}$48.8$ \\
        & OmniParser~\citep{lu2024omniparser} & $93.9$ & $57.0$ & $91.3$ & $63.6$ & $81.3$ & $51.0$ & \cellcolor{blue!10}$73.0$ \\
        & UGround-7B~\citep{gou2024navigating} & $90.1$ & $70.3$ & $87.1$ & $55.7$ & $85.7$ & $64.6$ & \cellcolor{blue!10}$75.6$ \\
        \midrule
        \multirow{2}{*}{GPT-4o~\citep{hurst2024gpt} } 
        & SeeClick~\citep{cheng2024seeclickharnessingguigrounding} & $81.0$ & $59.8$ & $69.6$ & $33.6$ & $43.9$ & $26.2$ & \cellcolor{blue!10}$52.3$ \\
        & UGround-7B~\citep{gou2024navigating} & $93.4$ & $76.9$ & $92.8$ & $67.9$ & $88.7$ & $68.9$ & \cellcolor{blue!10}$81.4$ \\
        \midrule
        \multicolumn{9}{l}{\textbf{Agent Model}} \\
    \midrule
        \multicolumn{2}{l}{GPT-4~\citep{openai2023gpt4}} & $22.6$ & $24.5$ & $20.2$ & $11.8$ & $9.2$ & $8.8$ & \cellcolor{blue!10}$16.2$ \\
        \multicolumn{2}{l}{GPT-4o~\citep{hurst2024gpt} } & $20.2$ & $24.9$ & $21.1$ & $23.6$ & $12.2$ & $7.8$ & \cellcolor{blue!10}$18.3$ \\
        \multicolumn{2}{l}{CogAgent~\citep{hong2024cogagent}} & $67.0$ & $24.0$ & $74.2$ & $20.0$ & $70.4$ & $28.6$ & \cellcolor{blue!10}$47.4$ \\
        \multicolumn{2}{l}{CogAgent-9B-20241220~\citep{hong2024cogagent}} & - & - & - & - & - & - & \cellcolor{blue!10}$85.4$ \\
        \multicolumn{2}{l}{SeeClick~\citep{cheng2024seeclickharnessingguigrounding}} & $78.0$ & $52.0$ & $72.2$ & $30.0$ & $55.7$ & $32.5$ & \cellcolor{blue!10}$53.4$ \\
        \multicolumn{2}{l}{Qwen2-VL~\citep{wang2024qwen2vlenhancingvisionlanguagemodels}} & $75.5$ & $60.7$ & $76.3$ & $54.3$ & $35.2$ & $25.7$ & \cellcolor{blue!10}$55.3$ \\
        \multicolumn{2}{l}{UGround-7B~\citep{gou2024navigating}} & $82.8$ & $60.3$ & $82.5$ & $63.6$ & $80.4$ & $70.4$ & \cellcolor{blue!10}$73.3$ \\
        % \multicolumn{2}{r}{UGround-2B} & $89.4$ & $72.0$ & $88.7$ & $65.7$ & $81.3$ & $68.9$ & $77.7$ \\
        % \multicolumn{2}{r}{UGround-7B} & $93.0$ & $79.9$ & $93.8$ & $76.4$ & $90.9$ & $84.0$ & $86.3$ \\
        \multicolumn{2}{l}{Aguvis-G-7B~\citep{xu2024aguvis}}& $88.3$ & $78.2$ & $88.1$ & $70.7$ & $85.7$ & $74.8$ & \cellcolor{blue!10}$81.8$ \\
        \multicolumn{2}{l}{OS-Atlas-7B~\citep{wu2024atlas}} & $93.0$ & $72.9$ & $91.8$ & $62.9$ & $90.9$ & $74.3$ & \cellcolor{blue!10}$82.5$ \\
        \multicolumn{2}{l}{Claude Computer Use~\citep{claude-computer-use}} & - & - & - & - & - & - & \cellcolor{blue!10}$83.0$ \\
        \multicolumn{2}{l}{Gemini 2.0 (Project Mariner)~\citep{geimini}} & - & - & - & - & - & - & \cellcolor{blue!10}$84.0$ \\
        \multicolumn{2}{l}{Aguvis-7B~\citep{xu2024aguvis}} & $\mathbf{95.6}$ & $77.7$ & $93.8$ & $67.1$ & $88.3$ & $75.2$ & \cellcolor{blue!10}$84.4$ \\
        \multicolumn{2}{l}{Aguvis-72B~\citep{xu2024aguvis}} & $94.5$ & $\mathbf{85.2}$ & $95.4$ &  $77.9$ & $\mathbf{91.3}$ & $\mathbf{85.9}$ & \cellcolor{blue!10}$89.2$ \\
        \midrule
        \multicolumn{2}{l}{\ourmodel-2B} & $93.0$ & $75.5$ & $90.7$ & $68.6$ & $84.3$ & $74.8$ & \cellcolor{blue!10}$82.3$ \\
        \multicolumn{2}{l}{\ourmodel-7B} & $94.5$ & $\mathbf{85.2}$ & $\mathbf{95.9}$ & $85.7$ & $90.0$ & $83.5$ & \cellcolor{blue!10}$\mathbf{89.5}$ \\
        \multicolumn{2}{l}{\ourmodel-72B} & $94.9$ & $82.5$ & $89.7$ & $\mathbf{88.6}$ & $88.7$ & $85.0$ & \cellcolor{blue!10}$88.4$ \\
        \bottomrule
    \end{tabular}
    }
    \caption{\small{Comparison of various planners and grounding methods on \ScreenSpot.}}
    \label{tab:screenspot_comparison}
\end{table}

\begin{table}[!t]
    \centering
     \resizebox{0.85\textwidth}{!}{%
    \begin{tabular}{llccccccccc}
        \toprule
        \multirow{2}{*}{\textbf{Method}} &  & \multicolumn{2}{c}{\textbf{Mobile}} & \multicolumn{2}{c}{\textbf{Desktop}} & \multicolumn{2}{c}{\textbf{Web}} & \multirow{2}{*}{\textbf{Avg}} \\
        \cmidrule(lr){3-4} \cmidrule(lr){5-6} \cmidrule(lr){7-8} % Partial horizontal line
        & & \textbf{Text} & \textbf{Icon/Widget} & \textbf{Text} & \textbf{Icon/Widget} & \textbf{Text} & \textbf{Icon/Widget} & \\
        \midrule
        \multicolumn{9}{l}{\textbf{Agent Framework}} \\
    \midrule
        \multirow{3}{*}{GPT-4o~\citep{hurst2024gpt} } 
        & SeeClick~\citep{cheng2024seeclickharnessingguigrounding} & $85.2$ & $58.8$ & $79.9$ & $37.1$ & $72.7$ & $30.1$ & \cellcolor{blue!10}$63.6$ \\
        & OS-Atlas-4B~\citep{wu2024atlas}    & $95.5$ & $75.8$ & $79.4$ & $49.3$ & $90.2$ & $66.5$ & \cellcolor{blue!10}$79.1$ \\
        & OS-Atlas-7B~\citep{wu2024atlas}    & $96.2$ & $83.4$ & $89.7$ & $69.3$ & $\mathbf{94.0}$ & $79.8$ & \cellcolor{blue!10}$87.1$ \\
        \midrule
        % \multirow{3}{*}{None} 
        \multicolumn{9}{l}{\textbf{Agent Model}} \\
    \midrule
        \multicolumn{2}{l}{SeeClick~\citep{cheng2024seeclickharnessingguigrounding}} & $78.4$ & $50.7$ & $70.1$ & $29.3$ & $55.2$ & $32.5$ & \cellcolor{blue!10}$55.1$ \\ 
        \multicolumn{2}{l}{OS-Atlas-4B~\citep{wu2024atlas}}   & $87.2$ & $59.7$ & $72.7$ & $46.4$ & $85.9$ & $63.1$ & \cellcolor{blue!10}$71.9$\\
        \multicolumn{2}{l}{OS-Atlas-7B~\citep{wu2024atlas}}  & $95.2$ & $75.8$ & $90.7$ & $63.6$ & $90.6$ & $77.3$ & \cellcolor{blue!10}$84.1$ \\
        \midrule
        % \multirow{2}{*}{None} 
        \multicolumn{2}{l}{\ourmodel-2B} & $95.2$ & $79.1$ & $90.7$ & $68.6$ & $87.2$ & $78.3$ & \cellcolor{blue!10}$84.7$ \\
        \multicolumn{2}{l}{\ourmodel-7B} & $\mathbf{96.9}$ & $\mathbf{89.1}$ & $\mathbf{95.4}$ & $85.0$ & $93.6$ & $85.2$ & \cellcolor{blue!10}$\mathbf{91.6}$ \\
        \multicolumn{2}{l}{\ourmodel-72B} & $94.8$ & $86.3$ & $91.2$ & $\mathbf{87.9}$ & $91.5$ & $\mathbf{87.7}$ & \cellcolor{blue!10}$90.3$ \\
        \bottomrule
    \end{tabular}
    }
    \caption{\small{Comparison of various planners and grounding methods on \ScreenSpotVTwo.}}
    \label{tab:screenspot_v2_comparison}
\end{table}

\ourmodel consistently outperforms baselines across multiple benchmarks. Specifically, in Table \ref{tab:screenspot_pro_comparison}, \ourmodel-72B achieves a score of $38.1$ on ScreenSpot Pro, significantly exceeding the performance of UGround-V1-7B ($31.1$) and OS-Atlas-7B ($18.9$). Notably, we observe that increasing the input image resolution on ScreenSpot Pro led to a significant performance improvement. Additionally, \ourmodel-7B attains a leading score of $89.5$ on ScreenSpot in Table \ref{tab:screenspot_comparison}. On ScreenSpot v2, as shown in Table \ref{tab:screenspot_v2_comparison}, both \ourmodel-7B ($91.6$) and \ourmodel-72B ($90.3$) outperform existing baselines, such as OS-Atlas-7B ($87.1$), further highlighting the robustness of our approach. In addition, the results show a significant improvement in grounding performance as we scale from \ourmodel-2B to \ourmodel-7B across all three grounding datasets. Comparing \ourmodel-7B and \ourmodel-72B, while ScreenSpot v1 and v2 exhibit no significant performance change, ScreenSpot Pro shows notable improvement in model scaling. This indicates that ScreenSpot v1 and v2 may not be sufficiently robust to fully capture the model's grounding capabilities at higher scales.

In summary, these results highlight the robust grounding capabilities of \ourmodel across various scenarios, including mobile, desktop, web, and professional environments. The models' consistent performance across datasets and their ability to handle both general and high-complexity tasks underscore their versatility and effectiveness in real-world GUI grounding applications.

\subsection{Offline Agent Capability Evaluation}
% Given that most multistep task datasets lack explicit thought processes or reasoning traces, we primarily assess the System 1 reasoning capabilities of agent models. System 1 reasoning refers to fast, intuitive decision-making without the explicit deliberation of System 2. These evaluations emphasize the models' ability to infer and execute actions effectively in offline, multistep scenarios.

To evaluate the GUI agent capabilities of \ourmodel in static, pre-defined
environments, we conduct evaluations on three benchmarks: \textbf{Multimodal Mind2Web}~\citep{zheng2024gpt} is designed to create and evaluate generalist web agents executing language instructions. 
It primarily assesses a model's performance in web-based environments. Metrics include element accuracy (Ele.Acc), operation F1 score (Op.F1), and step success rate (Step SR), as shown in Table~\ref{tab:mind2web}. \textbf{Android Control}~\citep{li2024on} evaluates planning and action-execution abilities in mobile environments. 
This dataset includes two types of tasks: (1) high-level tasks require the model to autonomously plan and execute multistep actions; (2) low-level tasks instruct the model to execute predefined, human-labeled actions for each step (Table~\ref{tab:mobile_tasks_results}). 
\textbf{GUI Odyssey}~\citep{lu2024gui} focuses on cross-app navigation tasks in mobile environments, featuring an average of 15+ steps per task. Tasks span diverse navigation scenarios with instructions generated from predefined templates. The dataset includes human demonstrations recorded on an Android emulator, providing detailed and validated metadata for each task episode. For Multimodal Mind2Web, we adhere to the settings and metrics specified in the original framework. For Android Control and GUI Odyssey (Table~\ref{tab:mobile_tasks_results}), we follow the settings and metrics outlined in OS-Atlas~\citep{wu2024atlas}.

\begin{table}[!t]
    \centering
    \resizebox{\textwidth}{!}{%
    \begin{threeparttable} % 关键部分
    \begin{tabular}{llccc ccc ccc}
    \toprule
    & \multirow{2}{*}{\textbf{Method}} & \multicolumn{3}{c}{\textbf{Cross-Task}} & \multicolumn{3}{c}{\textbf{Cross-Website}} & \multicolumn{3}{c}{\textbf{Cross-Domain}} \\
    \cmidrule(lr){3-5} \cmidrule(lr){6-8} \cmidrule(lr){9-11} % Partial horizontal line
     & & \textbf{Ele.Acc} & \textbf{Op.F1} & \textbf{Step SR} & \textbf{Ele.Acc} & \textbf{Op.F1} & \textbf{Step SR} & \textbf{Ele.Acc} & \textbf{Op.F1} & \textbf{Step SR} \\
    \midrule
    \multicolumn{11}{l}{\textbf{Agent Framework}} \\
    \midrule
     GPT-4o~\citep{hurst2024gpt}  & SeeClick~\citep{cheng2024seeclickharnessingguigrounding} & $32.1$ & - & - & $33.1$ & - & - & $33.5$ & - & - \\
    GPT-4o~\citep{hurst2024gpt}  & UGround~\citep{gou2024navigating} & $47.7$ & - & - & $46.0$ & - & - & $46.6$ & - & - \\ 
    GPT-4o~\citep{hurst2024gpt}  & Aria-UI~\citep{yang2024aria} & $57.6$ & - & - & $57.7$ & - & - & $61.4$ & - & - \\
    GPT-4V~\citep{gpt4v} & OmniParser~\citep{lu2024omniparser} & $42.4$ & $87.6$ & $39.4$ & $41.0$ & $84.8$ & $36.5$ & $45.5$ & $85.7$ & $42.0$ \\
    \midrule
    \multicolumn{11}{l}{\textbf{Agent Model}} \\
    \midrule
    \multicolumn{2}{l}{GPT-4o~\citep{hurst2024gpt} } & $5.7$ & $77.2$ & $4.3$ & $5.7$ & $79.0$ & $3.9$ & $5.5$ & $86.4$ & $4.5$ \\
    % \multicolumn{2}{r}{SeeClick} & $23.8$ & - & - & $15.3$ & - & - & $16.2$ & - & - \\
    \multicolumn{2}{l}{GPT-4(SOM)~\citep{achiam2023gpt}} & $29.6$ & - & $20.3$ & $20.1$ & - & $13.9$ & $27.0$ & - & $23.7$ \\ 
    \multicolumn{2}{l}{GPT-3.5(Text-only)~\citep{gpt35}} & $19.4$ & $59.2$ & $16.8$ & $14.9$ & $56.5$ & $14.1$ & $25.2$ & $57.9$ & $24.1$ \\
    \multicolumn{2}{l}{GPT-4(Text-only)~\citep{achiam2023gpt}} & $40.8$ & $63.1$ & $32.3$ & $30.2$ & $61.0$ & $27.0$ & $35.4$ & $61.9$ & $29.7$ \\ 
    % \midrule
    % \multicolumn{2}{r}{GPT-4(Text-only)} & $46.4$ & $73.4$ & $40.2$ & $38.0$ & $67.8$ & $32.4$ & $42.4$ & $69.3$ & $36.8$  \\
    \multicolumn{2}{l}{Claude\tnote{*}~\citep{claude-computer-use}} & $62.7$ & $84.7$ & $53.5$ & $59.5$ & $79.6$ & $47.7$ & $64.5$ & $85.4$ & $56.4$ \\
    \multicolumn{2}{l}{Aguvis-7B~\citep{xu2024aguvis}} & $64.2$ & $89.8$ & $60.4$ & $60.7$ & $88.1$ & $54.6$ & $60.4$ & $89.2$ & $56.6$ \\
    \multicolumn{2}{l}{CogAgent~\citep{hong2024cogagent}}& - & - & $62.3$ & - & - & $54$ & - & - & $59.4$ \\ 
    \multicolumn{2}{l}{Aguvis-72B~\citep{xu2024aguvis}} & $69.5$ & $90.8$ & $64.0$ & $62.6$ & $88.6$ & $56.5$ & $63.5$ & $88.5$ & $58.2$ \\
    \midrule
    \multicolumn{2}{l}{\ourmodel-2B} & $62.3$ & $90.0$ & $56.3$ & $58.5$ & $87.2$ & $50.8$ & $58.8$ & $89.6$ & $52.3$ \\
    \multicolumn{2}{l}{\ourmodel-7B} & $73.1$ & $92.2$ & $67.1$ & $68.2$ & $90.9$ & $61.7$ & $66.6$ & $90.9$ & $60.5$ \\
    \multicolumn{2}{l}{\ourmodel-72B} & $\mathbf{74.7}$ & $\mathbf{92.5}$ & $\mathbf{68.6}$ & $\mathbf{72.4}$ & $\mathbf{91.2}$ & $\mathbf{63.5}$ & $\mathbf{68.9}$ & $\mathbf{91.8}$ & $\mathbf{62.1}$ \\
    \bottomrule
    \end{tabular}
    \begin{tablenotes}
            \item[*] Claude refers to \textit{Claude-computer-use}.
        \end{tablenotes}
    \end{threeparttable}
    }
    \caption{\small{Performance comparison on Multimodal Mind2Web across different settings. We report element accuracy (Ele.Acc), operation F1 (Op.F1), and step success rate (Step SR).}}
    \label{tab:mind2web}
\end{table}

\begin{table}[!t]
    \centering
    \resizebox{0.85\textwidth}{!}{%
    \begin{threeparttable} % 关键部分
    \begin{tabular}{lccccccccc}
        \toprule
        \multirow{2}{*}{\textbf{Agent Models}} & \multicolumn{3}{c}{\textbf{AndroidControl-Low}} & \multicolumn{3}{c}{\textbf{AndroidControl-High}} & \multicolumn{3}{c}{\textbf{\GUIOdyssey}} \\
        \cmidrule(lr){2-4} \cmidrule(lr){5-7} \cmidrule(lr){8-10}
        & \textbf{Type} & \textbf{Grounding} & \textbf{SR} & \textbf{Type} & \textbf{Grounding} & \textbf{SR} & \textbf{Type} & \textbf{Grounding} & \textbf{SR} \\
        \midrule
        Claude\tnote{*}~\citep{claude-computer-use} & $74.3$ & $0.0$ & $19.4$ & $63.7$ & $0.0$ & $12.5$ & $60.9$ & $0.0$ & $3.1$ \\
        GPT-4o~\citep{hurst2024gpt}  & $74.3$ & $0.0$ & $19.4$ & $66.3$ & $0.0$ & $20.8$ & $34.3$ & $0.0$ & $3.3$ \\
        SeeClick~\citep{cheng2024seeclickharnessingguigrounding} & $93.0$ & $73.4$ & $75.0$ & $82.9$ & $62.9$ & $59.1$ & $71.0$ & $52.4$ & $53.9$ \\
        InternVL-2-4B~\citep{chen2024internvl} & $90.9$ & $84.1$ & $80.1$ & $84.1$ & $72.7$ & $66.7$ & $82.1$ & $55.5$ & $51.5$ \\
        Qwen2-VL-7B~\citep{wang2024qwen2vlenhancingvisionlanguagemodels} & $91.9$ & $86.5$ & $82.6$ & $83.8$ & $77.7$ & $69.7$ & $83.5$ & $65.9$ & $60.2$ \\
        Aria-UI~\citep{yang2024aria} & -- & $87.7$ & $67.3$ & -- & $43.2$ & $10.2$ & -- & $86.8$ & $36.5$ \\
        OS-Atlas-4B~\citep{wu2024atlas} & $91.9$ & $83.8$ & $80.6$ & $84.7$ & $73.8$ & $67.5$ & $83.5$ & $61.4$ & $56.4$ \\
        OS-Atlas-7B~\citep{wu2024atlas} & $93.6$ & $88.0$ & $85.2$ & $85.2$ & $78.5$ & $71.2$ & $84.5$ & $67.8$ & $62.0$ \\
        Aguvis-7B~\citep{xu2024aguvis} & -- & -- & $80.5$ & -- & -- & $61.5$ & -- & -- & -- \\
        Aguvis-72B~\citep{xu2024aguvis} & -- & -- & $84.4$ & -- & -- & $66.4$ & -- & -- & -- \\
        \midrule
        \ourmodel-2B & $\mathbf{98.1}$ & $87.3$ & $89.3$ & $81.2$ & $78.4$ & $68.9$ & $93.9$ & $86.8$ & $83.4$ \\
        \ourmodel-7B & $98.0$ & $89.3$ & $90.8$ & $83.7$ & $80.5$ & $72.5$ & $94.6$ & $90.1$ & $87.0$ \\
        \ourmodel-72B & $\mathbf{98.1}$ & $\mathbf{89.9}$ & $\mathbf{91.3}$ & $\mathbf{85.2}$ & $\mathbf{81.5}$ & $\mathbf{74.7}$ & $\mathbf{95.4}$ & $\mathbf{91.4}$ & $\mathbf{88.6}$ \\
        % \ourmodel-7B & $98.0$ & $89.3$ & $90.8$ & $83.7$ & $80.5$ & $72.5$ & $93.7$ & $-$ & $85.1$ \\
        % \ourmodel-72B & $\mathbf{98.1}$ & $\mathbf{89.9}$ & $\mathbf{91.3}$ & $\mathbf{85.2}$ & $\mathbf{81.5}$ & $\mathbf{74.7}$ & $\mathbf{94.4}$ & $-$ & $\mathbf{85.7}$ \\
        \bottomrule
    \end{tabular}
    \begin{tablenotes}
            \item[*] Claude refers to \textit{Claude-computer-use}.
        \end{tablenotes}
    \end{threeparttable}
    }
    \caption{\small{Results on mobile tasks (AndroidControl and GUI Odyssey). For AndroidControl, we report two settings (Low and High).}}
    \label{tab:mobile_tasks_results}
\end{table}

Across the three evaluated datasets, \ourmodel demonstrates clear advancements in reasoning and execution capabilities. In Multimodal Mind2Web (Table~\ref{tab:mind2web}), most of the agent models significantly outperform framework-based methods (using GPT-4o or GPT-4V as the core planner). Comparing different agent models, \ourmodel-72B achieving SOTA performance across key metrics. \ourmodel-7B, despite having fewer parameters, surpass strong baselines such as Aguvis-72B model and Claude. On AndroidControl and GUI Odyssey (Table~\ref{tab:mind2web}), \ourmodel-7B and \ourmodel-72B surpasses previous SOTA method (OS-Atlas-7B) with an absolute performance increase of $25$, showing notable superiority in multistep offline tasks. We also find that Claude Computer-Use performs strongly in web-based tasks but significantly struggles with mobile scenarios, indicating that the GUI operation ability of Claude has not been well transferred to the mobile domain. In contrast, \ourmodel exhibits excellent performance in both website and mobile domain, highlighting its adaptability and generalization capabilities. 

\begin{table}[!t]
    \centering
    \resizebox{0.7\textwidth}{!}{%
    \begin{threeparttable} % 关键部分
    \begin{tabular}{ll|cc}
        \toprule
        \multirow{2}{*}{\textbf{Method}} & &  \multicolumn{2}{c}{\textbf{Online}} \\
        \cmidrule(lr){3-4}
        & & \textbf{\OSWorld} & \textbf{\AndroidWorld} \\
        \midrule
        \multicolumn{4}{l}{\textbf{Agent Framework}} \\
         \midrule
         \multirow{5}{*}{GPT-4o~\citep{hurst2024gpt} } 
         & UGround~\citep{gou2024navigating} & - & $32.8$ \\
         & Aria-UI~\citep{yang2024aria} & $15.2$ & 44.8 \\
        & Aguvis-7B~\citep{xu2024aguvis} & $14.8$ & $37.1$ \\
         & Aguvis-72B~\citep{xu2024aguvis} & $17.0$ & - \\
         & OS-Atlas-7B~\citep{wu2024atlas} & $14.6$ & - \\
        \midrule
        \multicolumn{4}{l}{\textbf{Agent Model}} \\
        \midrule
         & GPT-4o~\citep{hurst2024gpt}  & $5.0$ & $34.5$ (SoM) \\
         & Gemini-Pro-1.5~\citep{team2024gemini} & $5.4$ & $22.8$ (SoM) \\
         & CogAgent-9B-20241220~\citep{hong2024cogagent} & $8.1$ & - \\
         % & Claude(15 steps)\tnote{*} & $14.9$ & - \\
         & Aguvis-72B~\citep{xu2024aguvis} & $10.3$ & $26.1$ \\
         % & Claude(50 steps)\tnote{*} & $22.0$ & - \\
         & Claude Computer-Use~\citep{claude-computer-use} & $14.9$ ($15$ steps) & $27.9$ \\
         & Claude Computer-Use~\citep{claude-computer-use} & $22.0$ ($50$ steps) & - \\
        \midrule
         & \ourmodel-7B-SFT & $17.7$ ($15$ steps) & $33.0$ \\
         & \ourmodel-7B-DPO & $18.7$ ($15$ steps) & - \\
         & \ourmodel-72B-SFT & $18.8$ ($15$ steps) & $\mathbf{46.6}$ \\
         & \ourmodel-72B-DPO & $22.7$ ($15$ steps) & - \\
         & \ourmodel-72B-DPO & $\mathbf{24.6}$ ($50$ steps) & - \\
        \bottomrule
    \end{tabular}
    \end{threeparttable}
    }
    \caption{\small{Results on online benchmarks. We evaluate performance under the screenshot-only setting on OSWorld, limiting the maximum number of steps to 15.}}
    \label{tab:online_evaluation}
\end{table}

\subsection{Online Agent Capability Evaluation}
\label{sec:exp_online}
Online evaluations enable dynamic environments, each designed as an interactive simulation that mirrors real-world scenarios. In these environments, GUI agents can alter environmental states by executing actions in real time. We evaluate different models in online environments using two benchmarks: \textbf{OSWorld}~\citep{xie2024osworldbenchmarkingmultimodalagents} provides a scalable and diverse environment for evaluating multimodal agents on complex tasks across Ubuntu, Windows, and macOS platforms. It consists of 369 tasks involving real-world web and desktop applications, with detailed setups and evaluation scripts. The evaluation is conducted in screenshot-only mode. To mitigate potential interference from network instability and environmental factors, the final score is averaged over 3 runs. We also consider traces where our model decides to ``CallUser'' or traces our model fails to output ``Finish'' in the end as infeasible tasks for evaluation.
\textbf{AndroidWorld}~\citep{rawles2024androidworld} is an environment designed for developing and benchmarking autonomous agents on a live Android emulator. It includes 116 tasks across 20 mobile apps, with dynamic task variations generated through randomized parameters. This dataset is ideal for evaluating agents' adaptability and planning abilities in mobile contexts.

The results are listed in Table~\ref{tab:online_evaluation}. (1) On \textbf{OSWorld}, when given a budget of $15$ steps, \ourmodel-7B-DPO ($18.7$) and \ourmodel-72B-DPO ($22.7$) significantly outperforms Claude ($14.9$), demonstrating its strong reasoning capabilities. Also, \ourmodel-72B-DPO with a $15$-step budget ($22.7$) is comparable to Claude when the latter is given $50$-step budget ($22.0$), showing great execution efficiency. Notably, \ourmodel-72B-DPO achieves a new SOTA result $24.6$ on OSWorld with a budget of $50$ steps, surpassing all the existing agent frameworks (e.g., GPT-4o with Aria-UI), highlighting the significant potential of agent models in addressing complex desktop-based tasks with higher efficiency and effectiveness.
(2) Results on \textbf{AndroidWorld} deliver a similar conclusion, with \ourmodel-72B-SFT achieving a $46.6$ performance, outperforming the best previous agent framework (GPT-4o with Aria-UI, $44.8$) and agent model (Aguvis-72B, $26.1$).
(3) Comparing the results of SFT model and DPO model, we find that DPO significantly improves the performance on OSWorld, showing that involving ``negative samples'' during training enables the model to better distinguish between optimal and suboptimal actions. (4) Furthermore, comparing \ourmodel-72B and \ourmodel-7B, we find that the 72B model performs much better than the 7B model in online tasks, and the gap is larger compared to offline tasks (Table~\ref{tab:mind2web} and Table~\ref{tab:mobile_tasks_results}). This shows that scaling model size significantly improves system 2 reasoning, enabling more deliberate and logical decision-making. 
Moreover, the discrepancy suggests that evaluations based solely on offline benchmarks may fail to accurately capture the models’ capabilities in real-time, dynamic environments.
In general, these results validate the potential of agent models for reasoning-intensive tasks and emphasize the advantages of leveraging larger-scale models to tackle the challenges of online environments.

\subsection{Comparing System 1 and System 2 Reasoning}
We compare the effects of system-1 and system-2 reasoning on model performance. System-1 reasoning refers to the model directly producing actions without chain-of-thought, while system-2 reasoning involves a more deliberate thinking process where the model generates reasoning steps before selecting an action. We train \ourmodel-7B to acquire both capabilities but we modify the model's reasoning behavior during inference through prompt engineering.

\paragraph{In-domain Evaluation}
We first evaluate performance across three in-domain agent benchmarks: Multimodal Mind2Web, Android Control, and GUI Odyssey, all of which have corresponding training data in \ourmodel. For efficiency in evaluation, we randomly sample 1,000 examples for the Android Control and GUI Odyssey benchmarks. We use the Best-of-N (BoN) sampling method, where \ourmodel samples $N$ candidate outputs per input, with $N$ set to 1, 16, and 64. The step success rate is used as the evaluation metric.

\begin{figure}[!t]
    \centering
    \begin{minipage}{0.98\textwidth}
        \centering
        \includegraphics[width=\textwidth]{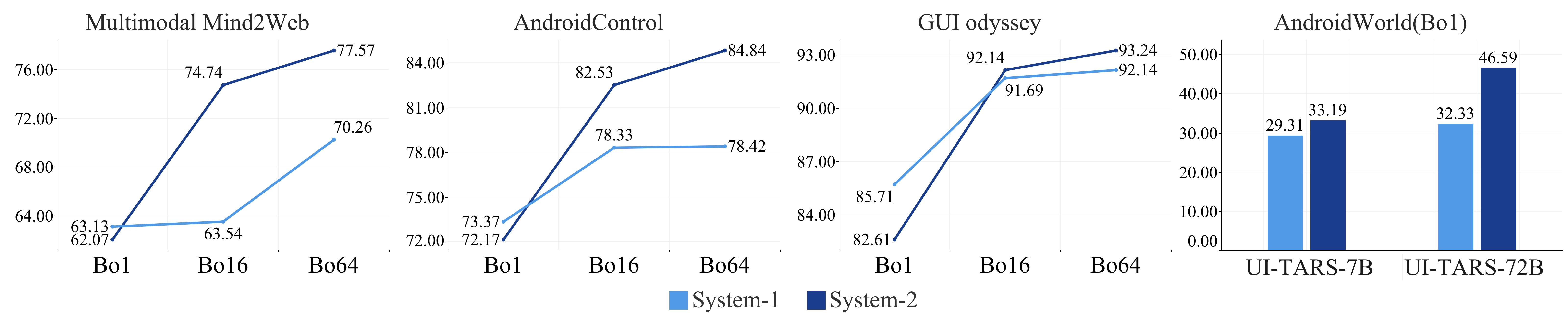}
        % \caption{MM2Web}
    \end{minipage}
    % \begin{minipage}{0.24\textwidth}
    %     \centering
    %     \includegraphics[width=\textwidth]{figures/AndroidControl_boN.jpeg}
    %     % \caption{Android Control}
    % \end{minipage}
    % \begin{minipage}{0.24\textwidth}
    %     \centering
    %     \includegraphics[width=\textwidth]{figures/GUIOdyssey_boN.jpeg}
    %     % \caption{GUI Odyssey}
    % \end{minipage}
    % % \caption{Performance comparison across different datasets.}
    % \begin{minipage}{0.24\textwidth}
    %     \centering
    %     \includegraphics[width=\textwidth]{figures/AndroidWorld_ablation.jpeg}
    %     % \caption{GUI Odyssey}
    % \end{minipage}
    \caption{\small{Performance of system-1 (no-thought) and system-2 (with thought) in in-domain (Mind2Web, AndroidControl, GUI Odyssey) and out-of-domain (AndroidWorld) benchmarks.}}
    \label{tab:boN_results}
\end{figure}

As shown in Figure~\ref{tab:boN_results}, at $N$=1, system-2 reasoning performs slightly worse than system-1 reasoning across all three in-domain benchmarks. While system-2 reasoning is generally expected to improve task execution by introducing a reflective, multi-step process, this result suggests that under a single-sample condition, the complexity of system-2 reasoning can lead to suboptimal reasoning steps. Specifically, the model may introduce irrelevant or incorrect reasoning steps—such as referring to non-existent objects or making erroneous inferences—which increases the risk of hallucinations or failure to generate the correct action. In the absence of diverse candidate outputs, the model may fixate on a flawed reasoning path, resulting in a lower likelihood of choosing the correct action.

However, as $N$ increases to 16 and 64, the system-2 model begins to demonstrate a clear advantage over system-1 reasoning. The increased number of candidate outputs provides greater diversity in the decision space, allowing the model to overcome suboptimal reasoning paths. In particular, the system-2 model benefits from the opportunity to explore multiple reasoning chains, which compensates for the earlier issues seen with $N$=1. The diversity of candidates increases the likelihood that the correct action is among the sampled outputs, even if some of the intermediate reasoning steps were not ideal. This shift in performance is particularly striking as it shows that the deliberate, multi-step reasoning of system-2 can effectively compensate for its initial disadvantages when sufficient candidate outputs are available.

A key insight is that while system-2 reasoning excels with sufficient diversity, achieving optimal performance with a single, decisive output (as in Bo1) remains a significant challenge. The ideal future direction involves leveraging system-2 reasoning's strengths in diverse, real-world scenarios while minimizing the need for multiple samples. This could be accomplished through techniques like reinforced fine-tuning~\citep{jaech2024openai}, which would guide the model to produce the correct action with high confidence in a single pass.

\paragraph{Out-of-domain Evaluation}
Next, we evaluate both reasoning methods on AndroidWorld, an out-of-domain (OOD) benchmark without corresponding training data in \ourmodel. We evaluate \ourmodel-7B and \ourmodel-72B at Bo1. Interestingly, the results from AndroidWorld reveal a significant shift compared to the in-domain benchmarks. While system-1 reasoning performs well in in-domain scenarios (Mind2Web, Android Control, and GUI Odyssey), system-2 reasoning significantly outperforms system-1 in the OOD setting (AndroidWorld). This suggests that although system-2 may face challenges in in-domain scenarios, particularly under single-sample conditions, its deeper reasoning capabilities provide a distinct advantage in OOD situations. In these cases, the increased reasoning depth helps the model generalize to previously unseen tasks, highlighting the broader applicability and potential of system-2 reasoning in real-world, diverse scenarios.
\section{Conclusion}
In this paper, we introduced \ourmodel, a native GUI agent model that integrates perception, action, reasoning, and memory into a scalable and adaptive framework. Achieving state-of-the-art performance on challenging benchmarks such as OSWorld, \ourmodel outperforms existing systems like Claude and GPT-4o. We presented several novel innovations, including enhanced perception, unified action modeling, system-2 reasoning, and iterative refinement using online traces, all of which enable the agent to effectively handle complex GUI tasks with minimal human oversight. We also reviewed the evolution path of GUI agents, from rule-based systems to adaptive native models. We segment the development process into key stages based on the degree of human intervention and generalization capabilities, emphasizing the transition from text-based methods to pure-vision, end-to-end agent models. We also explored the core capabilities of the native agent model, including perception, action, reasoning, and memory, which form the foundation for future advancements in GUI agents. Looking ahead, while native agents represent a significant leap forward, the future lies in the integration of active and lifelong learning, where agents autonomously drive their own learning through continuous, real-world interactions.

\section*{Acknowledgements}
We thank Ziqian Wei, and Tianyu Zhang for figure drawing, Yiheng Xu, Tao Yu, Wenqian Wang, Xiaobo Qin, Zhiyong Wu, and Yi Lin, Junyuan Qi, Zihao Wang, Jiecao Chen, Mu Qiao, Congwu Shen, Ruo Wang, Mingxuan Wang, Lin Yan, Renjie Zheng, Guanlin Liu, Yuwen Xiong for suggestions, and Faming Wu, Sihang Yuan, Ziyuan Zhao, Jie Tang, Zhaoyi An, Yiran Wang, Linlin Ao, Bairen Yi, Yanghua Peng, Lishu Luo, Zhi Zhang, Zehua Wang, Lingjun Liu for supporting this work.

\newpage
\bibliographystyle{delta_tuning}
\bibliography{custom}

\newpage
\appendix
\section{Case Study}
\label{sec:case_study}
\begin{figure}[!h]
    \centering
    \includegraphics[width=\linewidth]{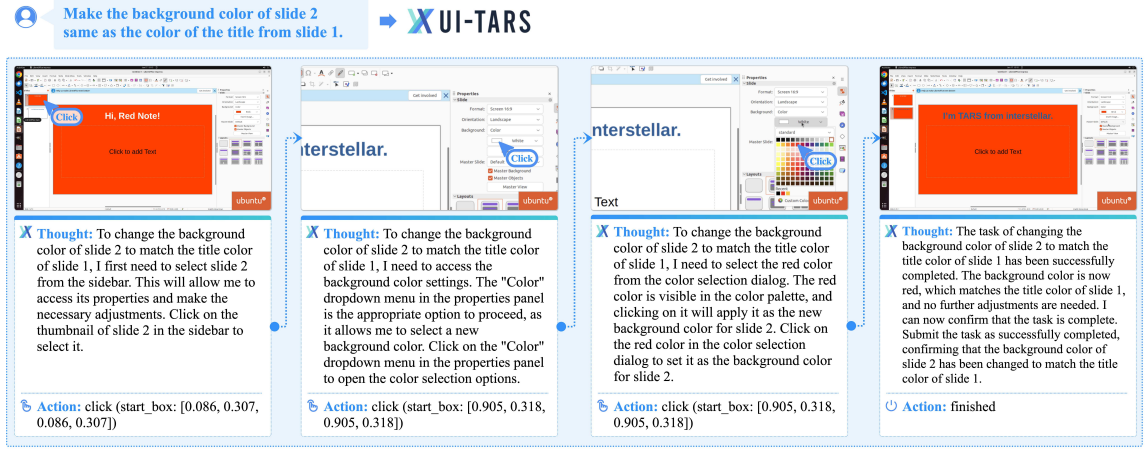}
    \caption{\small{Test case on Ubuntu impress scene from UI-TARS. The task is: Make the background color of slide 2 same as the color of the title from slide 1.}}
    \label{fig:case_test_2}
\end{figure}

\begin{figure}[!htbp]
    \centering
    \includegraphics[width=\linewidth]{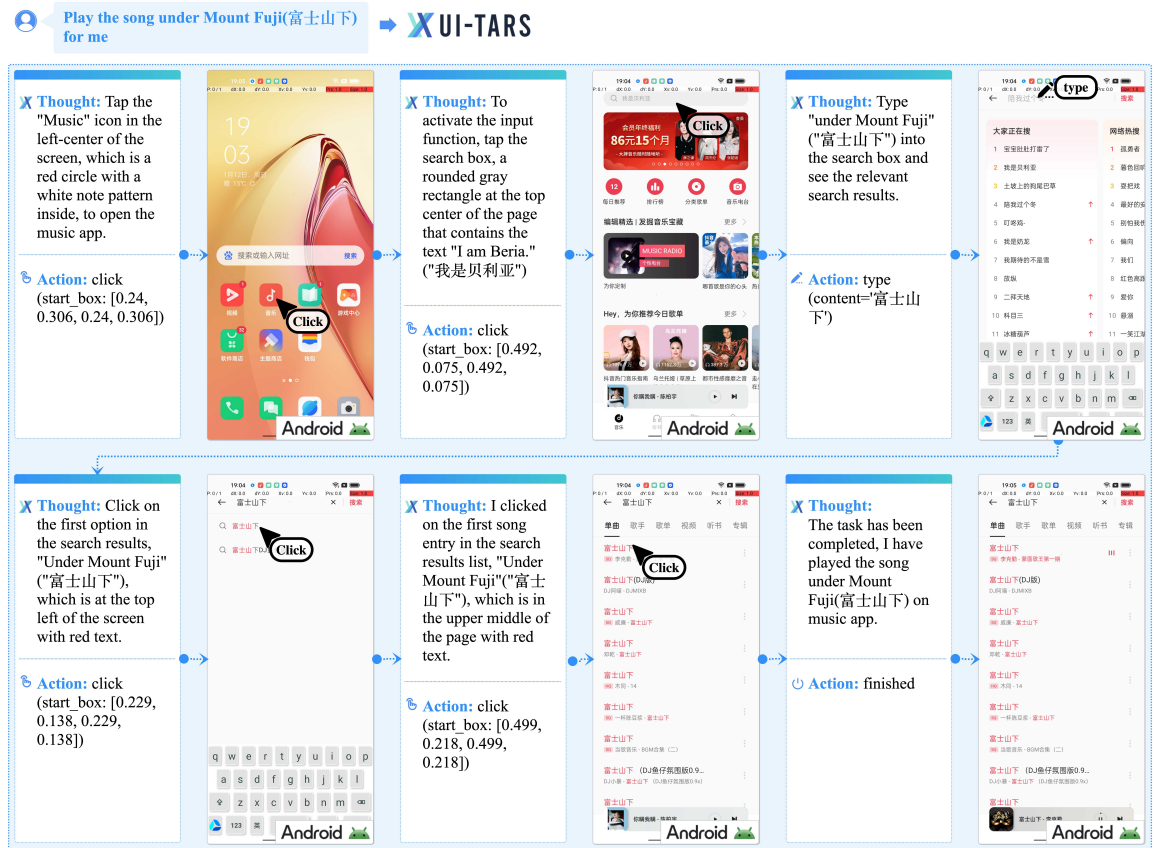}
    \caption{\small{Test case on Android from UI-TARS. The task is: Play the song under Mount Fuji.}}
    \label{fig:case_test_5}
\end{figure}

\begin{figure}[!t]
    \centering
    \includegraphics[width=\linewidth]{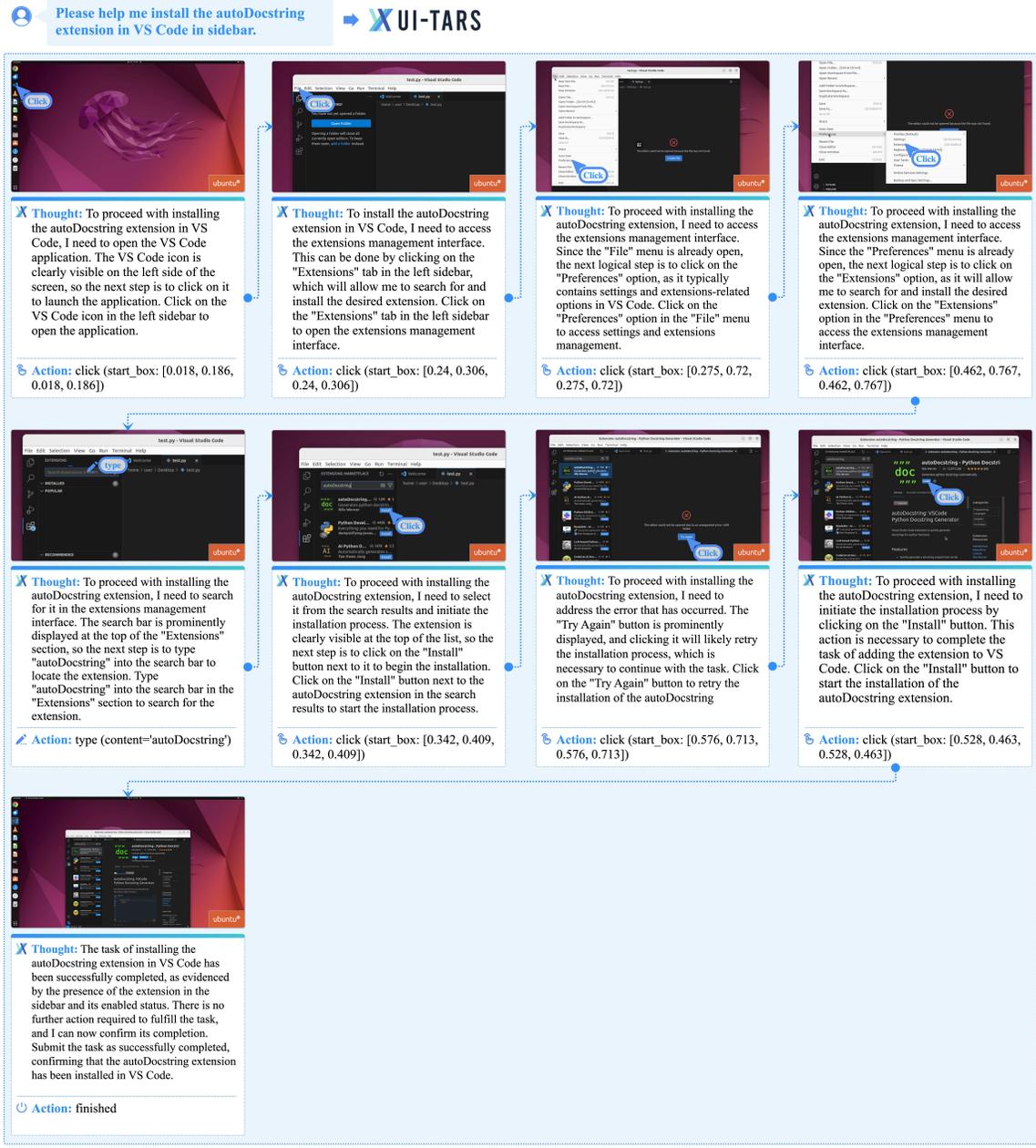}
    \caption{\small{Test case on Ubuntu VSCode scene from UI-TARS. The task is: Please help me install the autoDocstring extension in VS Code.}}
    \label{fig:case_test_3}
\end{figure}

\begin{figure}[!t]
    \centering
    \includegraphics[width=\linewidth]{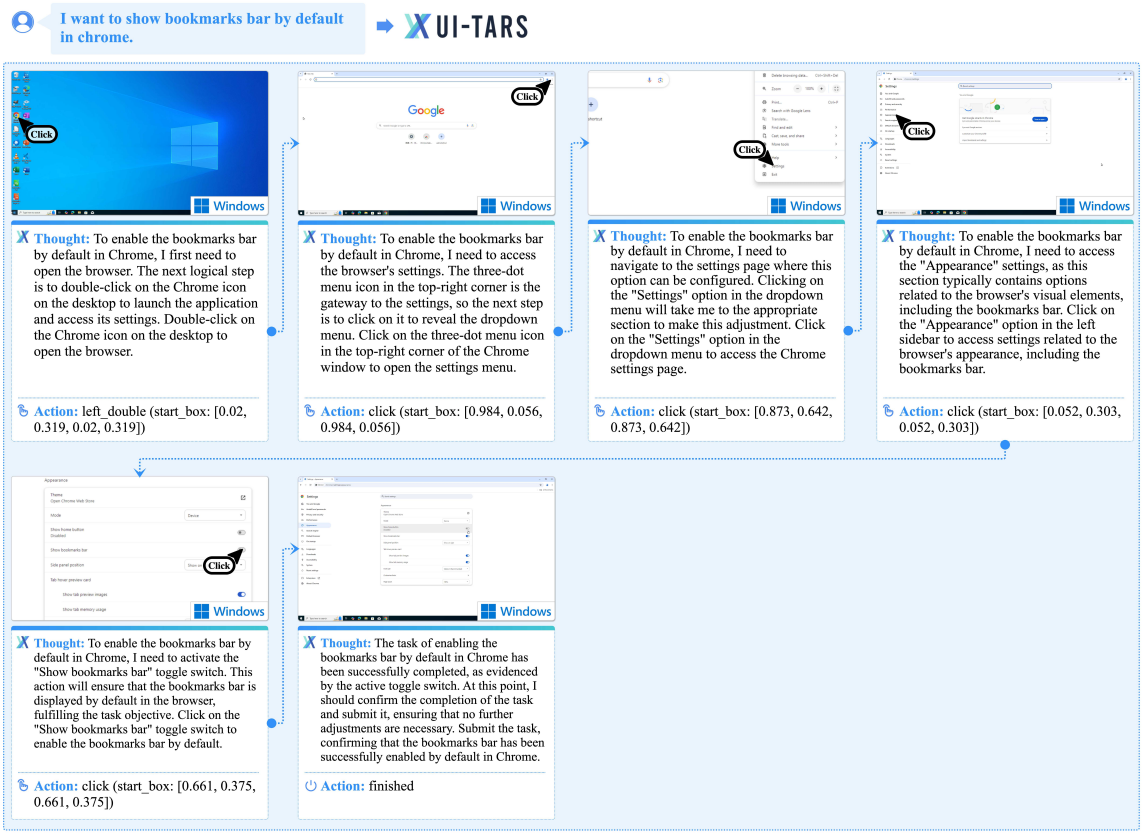}
    \caption{\small{Test case on Windows chrome scene from UI-TARS. The task is: I want to show bookmarks bar by default in chrome.}}
    \label{fig:case_test_4}
\end{figure}

We list a few cases executed by \ourmodel in Figure~\ref{fig:case_test_2}, Figure~\ref{fig:case_test_5}, Figure~\ref{fig:case_test_3}, and Figure~\ref{fig:case_test_4}.

\clearpage

\section{Data Example}
\label{sec:data_example}

\begin{figure}[!b]
    \centering
    \includegraphics[width=0.85\linewidth]{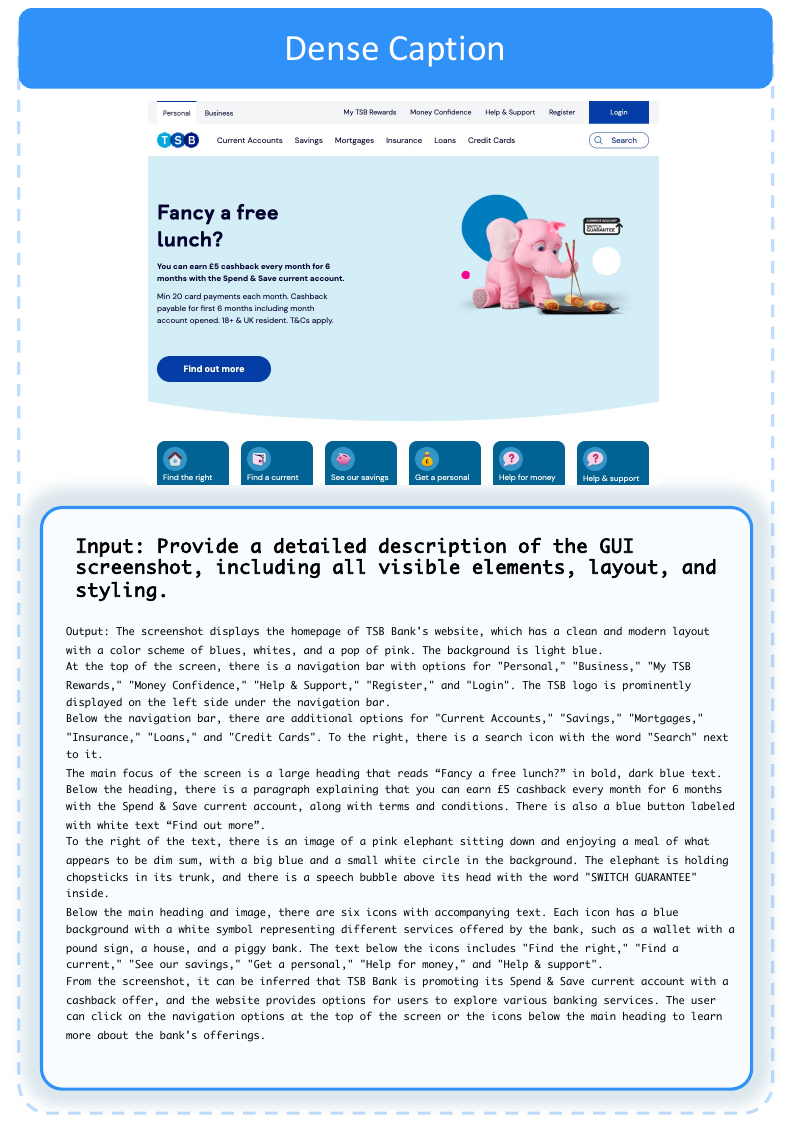}
    \caption{\small{Dense caption example.}}
    \label{fig:dense_caption}
\end{figure}

\begin{figure}[!t]
    \centering
    \includegraphics[width=0.95\linewidth]{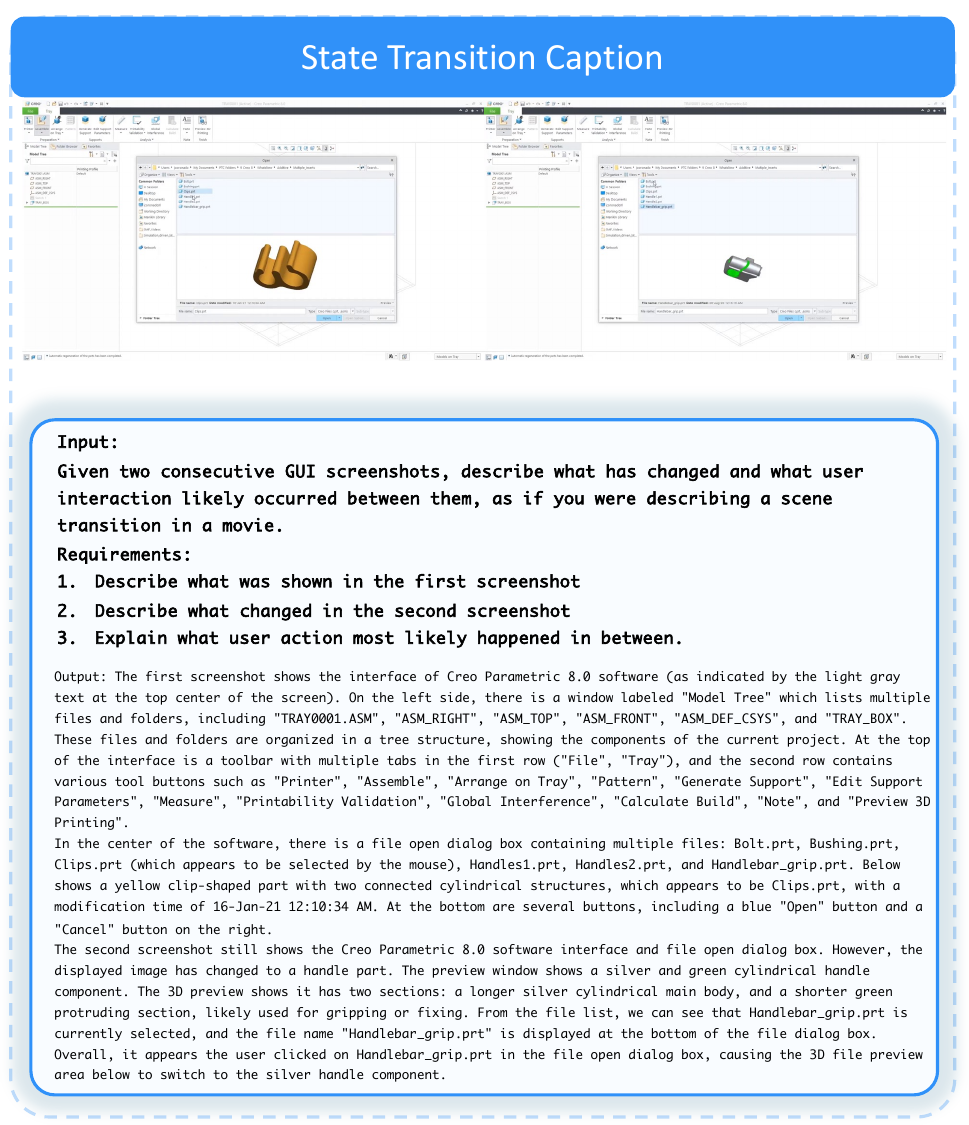}
    \caption{\small{State transition caption example.}}
    \label{fig:state_transition_caption}
\end{figure}

\begin{figure}[!t]
    \centering
    \includegraphics[width=0.95\linewidth]{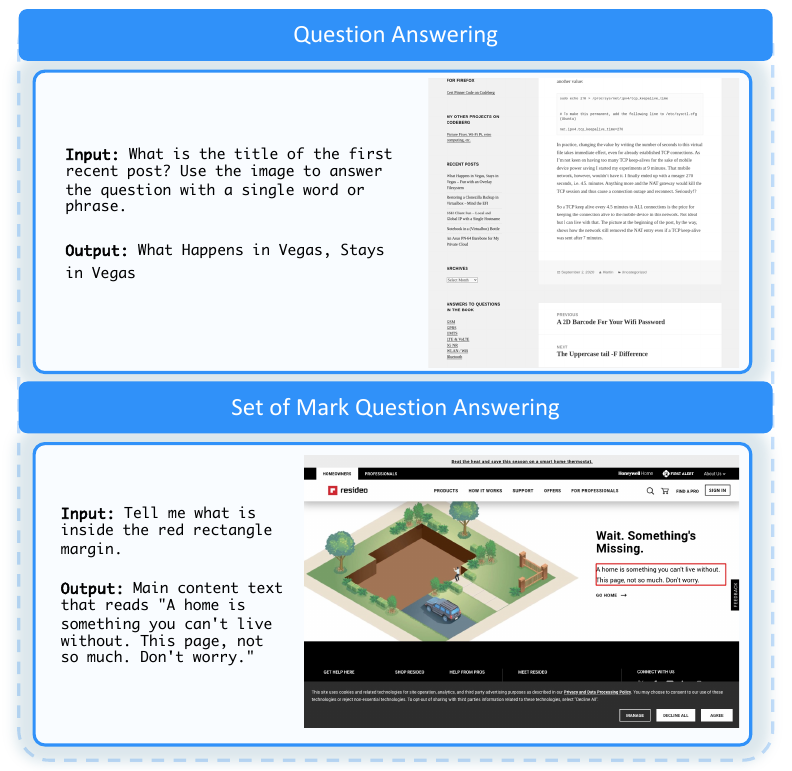}
    \caption{\small{Question answering and set of mark example.}}
    \label{fig:qa_and_som_qa}
\end{figure}

\begin{figure}[!t]
    \centering
    \includegraphics[width=0.95\linewidth]{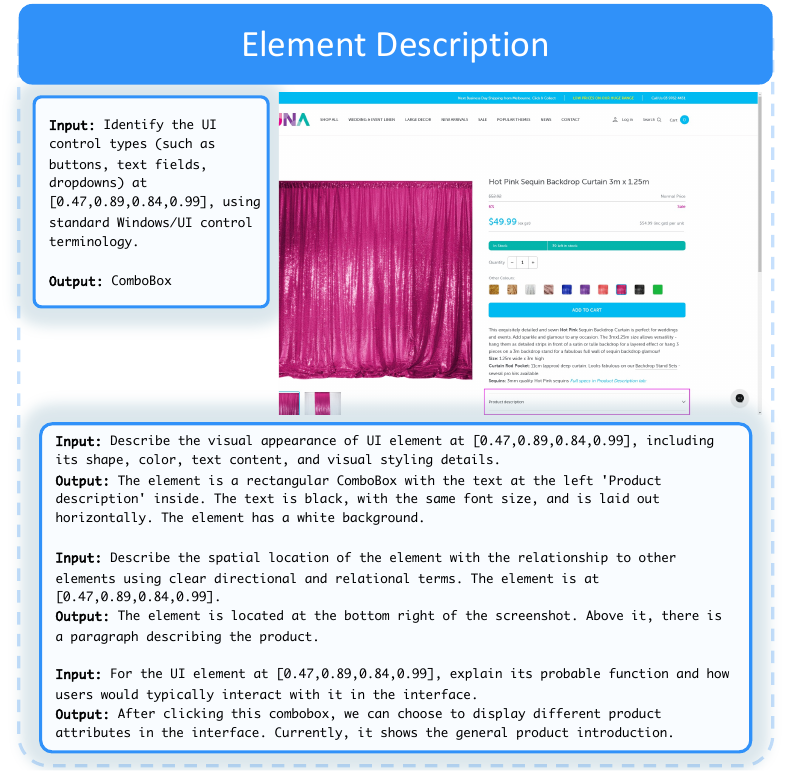}
    \caption{\small{Element description example, the target element is highlighted with a pink bounding box in the image.}}
    \label{fig:element_description}
\end{figure}

We show several data examples for perception training in Figure~\ref{fig:dense_caption}, Figure~\ref{fig:state_transition_caption}, Figure~\ref{fig:qa_and_som_qa}, and Figure~\ref{fig:element_description}. 

\end{document}